\documentclass[lettersize,journal]{IEEEtran}
\usepackage{amsmath,amsfonts}
\usepackage{algorithmic}
\usepackage{algorithm}
\usepackage{array}
\usepackage[caption=false,font=normalsize,labelfont=sf,textfont=sf]{subfig}
\usepackage{textcomp}
\usepackage{stfloats}
\usepackage{url}
\usepackage{verbatim}
\usepackage{graphicx}
\usepackage{cite}
\usepackage{caption}
\usepackage{float}
\usepackage{subfloat}
\usepackage[capitalize,noabbrev]{cleveref}

\usepackage{ulem}
\usepackage{color}
\usepackage{microtype}
\usepackage{makecell}
\usepackage{multirow} 

\usepackage{enumitem}
\usepackage{amssymb}
\usepackage{mathtools}
\usepackage{amsthm}

\usepackage{url}
\theoremstyle{plain}
\newtheorem{theorem}{Theorem}[section]

\theoremstyle{definition}

\theoremstyle{remark}

\begin{document}

\title{Efficient Machine Unlearning via Influence Approximation}

\author{Jiawei~Liu,
        Chenwang~Wu,
        Defu~Lian,
        and Enhong~Chen,~\IEEEmembership{Fellow,~IEEE}
\IEEEcompsocitemizethanks{\IEEEcompsocthanksitem J. Liu is with the School of Computer Science and Technology, University of Science and Technology of China, Hefei, Anhui 230000, China.\protect\\
E-mail: ljw1222@mail.ustc.edu.cn.
\IEEEcompsocthanksitem C. Wu is with the School of Artificial Intelligence and Data Science, University of Science and Technology of China, Hefei, Anhui 230000, China.\protect\\
E-mail: wcw1996@mail.ustc.edu.cn.
\IEEEcompsocthanksitem D. Lian and E. Chen are with the Anhui Province Key Laboratory of Big Data Analysis and Application, School of Computer Science and Technology, University of Science and Technology of China, Hefei, Anhui
230000, China.\protect\\
E-mail: \{liandefu, cheneh\}@ustc.edu.cn.
\IEEEcompsocthanksitem Corresponding author: Defu Lian.}}
%         <-this % stops a space



\maketitle

\begin{abstract}
Due to growing privacy concerns, machine unlearning, which aims at enabling machine learning models to ``forget" specific training data, has received increasing attention. Among existing methods, influence-based unlearning has emerged as a prominent approach due to its ability to estimate the impact of individual training samples on model parameters without retraining. 
However, this approach suffers from prohibitive computational overhead arising from the necessity to compute the Hessian matrix and its inverse across all training samples and parameters, rendering it impractical for large-scale models and scenarios involving frequent data deletion requests. This highlights the difficulty of forgetting. Inspired by cognitive science, which suggests that memorizing is easier than forgetting, this paper establishes a theoretical link between memorizing (incremental learning) and forgetting (unlearning). This connection allows machine unlearning to be addressed from the perspective of incremental learning. Unlike the time-consuming Hessian computations in unlearning (forgetting), incremental learning (memorizing) typically relies on more efficient gradient optimization, which supports the aforementioned cognitive theory.
Based on this connection, we introduce the Influence Approximation Unlearning (IAU) algorithm for efficient machine unlearning from the incremental perspective. Extensive empirical evaluations demonstrate that IAU achieves a superior balance among removal guarantee, unlearning efficiency, and comparable model utility, while outperforming state-of-the-art methods across diverse datasets and model architectures. Our code is available at \url{https://github.com/Lolo1222/IAU}.

\end{abstract}


\section{Impact statement}
Machine unlearning has traditionally been studied as a distinct research area, separate from the well-established field of incremental learning. While incremental learning has been extensively investigated for decades, machine unlearning has only recently gained attention due to growing concerns about data privacy and regulatory requirements. This paper makes a significant conceptual leap by establishing, for the first time, a theoretical connection between these two fields. Our work bridges this critical gap, enabling the transfer of robust methodologies and algorithmic insights from incremental learning to the design of efficient machine unlearning solutions.Beyond theoretical contributions, we present the Influence Approximation Unlearning (IAU) algorithm, which empirically validates the feasibility. By leveraging principles from incremental learning, IAU achieves superior performance compared to existing unlearning methods, offering a scalable and practical solution for privacy-aware machine learning systems. This advancement not only enriches the understanding of both fields but also opens new avenues for developing high-performance unlearning algorithms grounded in well-studied incremental learning techniques. Our work thus represents a substantial step forward in the intersection of machine unlearning, data privacy, and algorithmic innovation.

\section{Introduction}
\IEEEPARstart{I}{n} our daily lives, people generate vast amounts of data through social media updates, banking transactions, and cloud-synced location data. Organizations exploit user data to train personalized machine learning (ML) models. Machine unlearning \cite{towards}, an important field of ML research, has received increasing attention. It aims to erase sample-specific information from models efficiently. On the one hand, growing privacy concerns drive users to selectively delete sensitive information, aligning with privacy regulations such as the European Union's GDPR \cite{gdpr}, which grants individuals \textit{the right to be forgotten}, i.e., the right of an individual’s data to be deleted from a database (and derived products) and requires companies to delete personal data upon request. On the other hand, efficient machine unlearning is crucial in various scenarios, including removing contaminated data points due to data poisoning attacks \cite{rubinstein2009antidote,zhang2022poison}, eliminating outdated information \cite{wang2022causal}, and handling misleading or ambiguous data \cite{pang2021recorrupted}. These diverse needs underscore the importance of developing algorithms that enable models to quickly forget specific training points without significant utility loss.

An intuitive approach to implementing unlearning is to retrain the model from scratch based on the remaining data upon receiving a forgetting request. Although this can provide precise removal guarantees and maintain model utility, it is time-consuming and computationally expensive, especially when dealing with large-scale datasets and frequent forgetting requests. Therefore, it is crucial to design an efficient unlearning mechanism that balances the trade-offs among removal guarantee, unlearning efficiency, and comparable model utility.

Current unlearning approaches can be broadly categorized into two main classes: (1) \textbf{Exact unlearning} \cite{sisa}, which typically involves partitioning the training dataset into distinct shards, with each shard used to train an isolated sub-model. During inference, the results from all sub-models are aggregated to produce the final output.  When a forgetting request arrives, only the shard containing the data points to be forgotten is retrained, while other sub-models remain unchanged\cite{towards,sisa,ginart2019making,liu2022right}. This approach ensures precise removal of the targeted data but disrupts the inherent relationships between data points, leading to significant performance degradation. Additionally, retraining the affected sub-model, even for a single shard, \textit{remains a computationally expensive and time-consuming process}.
(2) \textbf{Approximate unlearning} \cite{guo2019certified}, which adjusts model parameters to scrub the contribution of unlearning data, ensuring approximate indistinguishability between the unlearned model and a retrained model \cite{sekhari2021remember,wu2022puma,ullah2021machine,unrolling,amne,fisher,lcodec,neel2021descent,ma2022learn}.
The influence function \cite{influence} shows great potential by using a first-order Taylor expansion of the loss function to estimate the effect of removing a single sample on model's parameters. While this aligns well with the objectives of machine unlearning and achieves a notable enhancement in model performance, the computation of influence function necessitates the calculation of the Hessian matrix across all model parameters and the entire dataset, followed by the subsequent inversion of this high-dimensional matrix. This procedure is inherently computationally intensive, as \textit{it requires significant computational resources and storage capacity to handle the large-scale matrix operations involved}.

The above discussion of existing methods highlights the challenges of unlearning, which often entails substantial computational demands and memory requirements to achieve data removal. Inspired by cognitive science, which suggests that memorizing is easier than forgetting \cite{wang2019more}, this paper attempts to address machine unlearning from the perspective of memory (incremental learning \cite{polikar2001learn}). 
Specifically, we first establish a theoretical bridge between incremental learning and machine unlearning. This theoretical breakthrough connects two previously distinct research domains and reveals a new perspective for implementing unlearning through incremental learning. Incremental learning typically relies on gradient-based optimization, which is more efficient than the time-consuming computation and inversion of Hessian matrices required for unlearning, thereby supporting the notion that ``memorizing is easier than forgetting." Furthermore, we propose a novel unlearning algorithm called \textbf{I}nfluence \textbf{A}pproximate \textbf{U}nlearning (IAU) that synergistically integrates incremental learning algorithms. IAU consists of three core modules: incremental approximation, gradient correction, and gradient restriction. Incremental approximation achieves the forgetting effect by incrementally learning negative samples of forgotten points, avoiding the need for costly Hessian matrix calculations and inversions. However, the gradient-based update strategy in incremental approximation may result in ``over-forget" and be affected by abnormal gradients. To address this, gradient correction in the unlearning phase adjusts gradient information for the remaining data, while gradient restriction during model training limits gradient size to mitigate the impact of abnormal gradients on unlearning updates.
Extensive experimental results prove that our IAU algorithm effectively balances multiple unlearning properties and delivers superior performance in comparison with state-of-the-art methods. 

Our main contributions are as follows:
\begin{itemize}[leftmargin = 10 pt]
% \begin{itemize}[leftmargin = 10 pt]
    \item Inspired by cognitive science that memorizing is easier than forgetting, we establish a bridge between incremental learning and machine unlearning through theoretical analysis and innovatively transform unlearning (forgetting) into incremental learning (memorizing).
    \item We propose IAU, a novel efficient unlearning framework developed under the perspective of incremental learning. This approach not only significantly reduces computational overhead and memory consumption but also demonstrates the potential of leveraging incremental learning methodologies to design unlearning mechanisms, thereby establishing a novel paradigm for advancing future algorithmic advancements in unlearning research. 
    % \item We pioneered the use of incremental learning to implement machine unlearning, which provides a new avenue for unlearning learning and may also provide inspiration for future work.
    \item We conducted comprehensive experiments to assess the efficacy of the proposed algorithm. The empirical results consistently demonstrate that the proposed IAU framework achieves a superior trade-off among removal guarantee, comparable model utility, and unlearning efficiency, outperforming existing state-of-the-art methods.
\end{itemize}

\section{Related Work}
\label{related}
\textbf{Machine unlearning} refers to the process of removing the influence of specific training data subsets from a trained model without necessitating full retraining. Driven by growing regulatory mandates and ethical imperatives surrounding data privacy—such as the General Data Protection Regulation (GDPR) \cite{gdpr}, the California Consumer Privacy Act of 2018 (CCPA) \cite{de2018guide}, and the UK Information Commissioner's Office (ICO) guidelines \cite{ico2020guidance}—this domain has emerged as a critical research area in machine learning. While retraining the model from scratch constitutes a naive solution that ensures complete data exclusion, this approach incurs prohibitive computational and temporal costs, rendering it impractical for large-scale systems \cite{towards,sisa}. To address this issue, two alternative forgetting routes have been proposed: exact unlearning and approximate unlearning.

\noindent\textbf{Exact unlearning} seeks to construct a new model that exactly performs the behavior of a model retrained on the remaining training data after specified samples are excluded. For foundational machine learning models, prior studies have explored unlearning techniques for specific algorithms: Ginart et al. \cite{ginart2019making} proposed k-means clustering-based approaches, while SVM-based unlearning methods were developed by Romero et al. \cite{romero2007incremental} and Karasuyama et al. \cite{karasuyama2009multiple}. Naive Bayes classifiers were also addressed in \cite{towards}. In the context of deep learning, the SISA framework \cite{sisa} employs a ``divide and conquer'' strategy by partitioning the training data into disjoint shards. Each subset is used to train isolated sub-models, which are subsequently aggregated into a consolidated final model. When handling unlearning requests, only the sub-models associated with the affected data shards are retrained. However, this partitioning approach introduces critical limitations: disjoint data shards disrupt inherent sample correlations within the dataset, leading to model performance degradation. Additionally, even when retraining a single data shard, the computational burden remains substantial, particularly under large-scale datasets and frequent unlearning operations. In such scenarios, the iterative retraining of sub-models incurs prohibitive time and resource costs, undermining the feasibility of SISA’s ``divide and conquer'' strategy in real-world applications where privacy compliance demands quick and efficient data exclusion.
    
% \noindent\textbf{Approximate unlearning} ensures nearly indistinguishable distributions in the final activation results between the unlearned and retrained models \cite{sekhari2021remember,golatkar2020forgetting}. 
% Influence-based unlearning, which is model-agnostic and minimally affects model utility, holds promise as an effective approach for achieving machine unlearning.
% Certified removal \cite{guo2019certified} first proposed to unlearn training data in linear models by removing its influence computed by influence function on model parameters. But computing influence for large neural networks requires the calculation of the exact inverse-Hessian Vector product, which is computationally expensive. LCODEC \cite{lcodec} employs an approximation method for the Hessian matrix, selecting a subset of parameters for updating to save time. Nevertheless, this approach often results in a significant gap between the accuracy achieved by applying its unlearning algorithm and the accuracy attained through retraining. For example, the accuracy will lower more than 10\% after 0.5\% removals for an MNIST Logistic Regressor. Therefore it is necessary to devise new influence-based methods that not only save time but also preserve the benefits of influence unlearning without harming the model utility.
\noindent\textbf{Approximate unlearning} aims to ensure that the unlearned model remain nearly indistinguishable from those of a model retrained on the remaining dataset \cite{sekhari2021remember,golatkar2020forgetting}. Among existing approaches, influence-based unlearning stands out as a promising model-agnostic method due to its minimal impact on model utility. The Certified Removal \cite{guo2019certified} pioneered this direction by unlearning linear models through influence function, which compute the parameter adjustments required to remove a specific training instance’s influence. However, when scaling this approach to large neural networks, the need to compute the Hessian matrix and inverse-Hessian vector products introduces prohibitively expensive computational demands. To address this, LCODEC \cite{lcodec} updates only a subset of parameters to reduce computational overhead. Nevertheless, the method exhibits a significant performance gap compared to retraining: for instance, unlearning 0.5\% of the training data in an MNIST logistic regressor results in over a 10\% accuracy drop. This trade-off underscores the urgent need for novel approximate unlearning techniques that simultaneously achieve computational efficiency and preserve model utility.
Our proposed method shares the same objective as approximate unlearning in preserving model utility while removing training data influence, but it is fundamentally distinguished by its novel conceptual framework rooted in incremental learning paradigms. Our approach innovatively redefines the unlearning process as a memorizing task, effectively transforming the conventional notion of ``forgetting" into  ``remembering". This paradigm shift offers a theoretically distinct solution in the unlearning problem domain.
% \textcolor{blue}{\noindent\textbf{Incremental learning} aims to}
\section{Preliminary}
This section will give necessary preliminaries, including the definition of machine unlearning and the influence function. Table \ref{tab: notation} shows the key notations used in the paper.

\begin{table}[tbp]
\small
	\centering
	\caption{Summary of key notations}
	\small
	\setlength{\tabcolsep}{0.004\linewidth}{
	\begin{tabular}{cc}
		\hline\noalign{\smallskip}
		Notation & Definition \\
		\noalign{\smallskip}\hline\noalign{\smallskip}
		$D$     & Training dataset \\
        $D_f$     & The data that the model should forget \\
        $D_r$     & The data that the model should remember \\
        $z_i$     & \makecell[c]{The $i$-th sample pair in $D$, which is associated\\ with a data $x_i$ and a label $y_i$}\\
        $z_-$ & A sample that needs to be removed\\
        $z_+$ & \makecell[c]{A sample added to the dataset $D$, which\\ can achieve the effect of unlearning $z_-$}\\
        $h(D)$ & The model $h$ is trained on the dataset $D$\\
        $h_u(h(D),D_f)$ & \makecell[c]{The sanitized model $h_u$ who approximates to\\ the model $h$ trained on $D-D_f$}\\
        $l(z,\theta)$ & Loss on $z$ for the model parameterized as $\theta$\\
		\noalign{\smallskip}\hline
	\end{tabular}%
}
	\label{tab: notation}%
\end{table}%

\subsection{Machine Unlearning}
\label{sec: machine unlearning}
Machine unlearning is a technique that enables trained models to forget previously learned data. This technique involves a training dataset of $N$ samples $D=\left\{z_i: (x_i, y_i)\right\}_{i=1}^N$, where each sample pair $z_i$ is associated with a data $x_i \in \mathbb{R}^d$ and a label $y_i \in \mathcal{Y}=\{1,2, \ldots, Y\}$, where $Y$ is the number of classes. A classification model $h(D)$ is trained on the complete training dataset $D$.

Users can submit a data removal request at any time, which partitions the dataset $D$ into two subsets: $D_f \subseteq D$, which represents the data that the model should forget, and  $D_r \subseteq D$, which represents the data that the model should remember. The goal of machine unlearning is to eliminate the influence of $D_f$ from $h(D)$. 

One solution is to use $D_r$ as the training data to retrain a new classification model $h(D_r)$ from scratch. However, this method can be time-consuming for large-scale datasets. A more efficient method is to use the unlearning mechanism $h_u$ to generate a sanitized model $h_u(h(D), D_f )$ directly from the deployed model $h(D)$, and we expect the unlearned model $h_u(h(D),D_f)$ is as similar to the retrained model $h(D_r)$ as possible, i.e.,
$$
h_u(h(D),D_f) \approx h(D_r).
$$

The measure of similarity is too broad. For this reason, there are multiple ideal properties that a good machine unlearning method must satisfy \cite{wu2023gif}:
\begin{itemize}[leftmargin = 10 pt]
\item \textbf{Removal Guarantee.}
The unlearning mechanism must completely remove the information of deleted data from the trained model, including the deleted data itself and its influences on other samples.
\item \textbf{Unlearning Efficiency.}
The unlearning mechanism should be time-efficient compared to model retraining.
\item \textbf{Comparable Model Utility.}
The unlearning mechanism should result in only a small utility gap compared to retraining from scratch in order to be practical.
\end{itemize}

\subsection{Influence Function}
\label{sec: inf}
Consider $h$ to be a function parameterized by $\theta$, which maps from an input feature space $\mathcal{X}$ to an output space denoted by $\mathcal{Y}$. The training samples are represented by the set $D=\left\{z_i:(x_i, y_i)\right\}_{i=1}^n$, while for a particular training sample $z$, the loss function is denoted as $\ell(z, \theta, h)$ (abbreviated as $\ell(z,\theta)$). The standard empirical risk minimization aims to solve the following optimization problem:
$$
\theta^*=\arg \min _\theta \frac{1}{n} \sum_{i=1}^n \ell(z_i,\theta) .
$$
% We assume $\ell$ is twice-differentiable and strictly convex in $\theta$, i.e.,
% $$H_{\theta^*}=\frac{1}{n}\sum_{i=1}^n\nabla_{\theta}^2\ell(z_i,\theta^*).$$
If a training example $z$ is up-weighted by an infinitesimal amount $\epsilon$, it results in a modified set of model parameters denoted as $\theta_{{z}}^\epsilon$. This modification is obtained by solving:
$$
\theta_{\{z\}}^{\epsilon *}=\arg \min _\theta \frac{1}{n} \sum_{i=1}^n \ell(z_i,\theta)+\epsilon \ell(z,\theta) .
$$
An intuitive approach is to retrain the entire model to obtain accurate $\theta_{\{z\}}^{\epsilon*}$, but as emphasized above, the time cost is intolerable. To this end, \cite{influence} suggested approximating $\theta_{\{z\}}^{\epsilon*}$ using the first-order Taylor series expansion around the optimal model parameters represented by $\theta^*$. 
This approximation yields:
$$
\theta_{\{z\}}^{\epsilon *} \approx \theta^*-\epsilon H_{\theta^*}^{-1} \nabla_\theta \ell(z,\theta^*).
$$
Here, $H_{\theta^*}$ represents the Hessian matrix with respect to the model parameters $\theta^*$, that is,
$$H_{\theta^*}=\frac{1}{n}\sum_{i=1}^n\nabla_{\theta}^2\ell(z_i,\theta^*).$$
As removing a point $z_-$ is equivalent to upweighting it by $\epsilon=-\frac{1}{n}$, we can approximate the changes in model parameters without having to retrain the model: 
\begin{equation}
    {\theta}^*_{\{z_-\}}-{\theta}^* \approx \frac{1}{n}H_{\theta^*}^{-1} \nabla_\theta \ell(z_-,\theta^*). 
    \label{eq: origin_inf}
\end{equation}
Here ${\theta}^*_{\{z_-\}}$ denotes the new empirical risk minimizer on dataset $D_r=D-\{z_-\}$.
In this work, we refer to it as ``influence unlearning''.

% \textcolor{blue}{Notes: Some works calculate removal influence as $-\frac{1}{n}H_{\theta^*}'^{-1} \nabla_{\theta} \ell(z_-,\theta^*)$ with $H_{\theta^*}'=\frac{1}{n-1}\sum_{z_i \in D_r}\nabla_{\theta}^2\ell(z_i,\theta^*)$. The difference lies in calculating the Hessian matrix on the origin training set or retain training set. They are both widely used in machine unlearning areas. For convenience, we use the form calculated on the origin training set. We further discuss the two forms of influence unlearning in Appendix xxx.}
\section{Methodology}

\subsection{Incremental Approximation}\label{subsection{Hessen matrix approximation}}
According to the major theory in cognitive science, it is often easier to memorize than to forget \cite{wang2019more}. Therefore, our study's starting point lies in a counterfactual: Can we approximately counteract the influence of unlearning points by adding a point? That is, approximating the forgetting effect through incremental learning of the original model. 
% \footnotemark\footnotetext{Note that we only require approximate unlearning, as complete unlearning without retraining cannot be achieved \cite{guo2020certified,liu2023certified}}.

Before answering this question, we first introduce a theorem about how much incremental learning of a point will cause the model's predictions to change.

\begin{theorem}[Influence of adding a point]\label{{theorem}[influence of adding a point]}
For a point $z$ and parameters $\theta$, let $ \ell (z, \theta)$ be the loss, and the empirical risk minimizer is given by $ {\theta}^* \stackrel{\text { def }}{=} \operatorname{argmin}_{\theta} \frac{1}{n} \sum_{i=1}^{n}
{\ell}({z}_{{i}}, \theta)$. 
We add a point $z_{+}$ to the training dataset. The new empirical risk minimizer is given by $ {\theta}^*_{\{z_+\}} \stackrel{\text { def }}{=} \operatorname{argmin}_{\theta} \frac{1}{{n+1}} (\sum_{{i}=1}^{{n}} {\ell}({z}_{{i}}, \theta)+{\ell}(z_+,\theta)). $
We have
$${\theta}^*_{\{z_+\}}-{\theta}^*\approx-\frac{1}{n} H_{{\theta}^*}^{-1} \nabla_\theta {\ell}({z}_{+}, {\theta}^*),
$$
% where ${H}_{{\theta}^*} \stackrel{\text { def }}{=} \nabla_\theta^2 {R}({\theta}^*)=\frac{1}{{n}} \sum_{{i}=1}^{{n}} \nabla_\theta^2 {\ell}({z}_{{i}}, \theta^*)$ is the Hessian matrix.
\end{theorem}
The proof of Theorem \ref{{theorem}[influence of adding a point]} can be found in Appendix \ref{proof}. Summarizing Eq. \ref{eq: origin_inf} and Theorem \ref{{theorem}[influence of adding a point]}, we can get the model parameters after adding a sample $z_+$ and deleting a sample $z_-$ respectively:
\begin{equation}
\begin{split}
    {\theta}^*_{\{z_+\}}\approx {\theta}^*-\frac{1}{n} H_{{\theta}^*}^{-1} \nabla_\theta {\ell}({z}_{+}, {\theta}^*)\\
    {\theta}^*_{\{z_-\}} \approx {\theta}^*+ \frac{{1}}{{n}} {H}_{{\theta}^*}^{-1} \nabla_\theta {\ell}({z}_{-}, {\theta}^*).
    \label{eq: parameters}
\end{split}
\end{equation}

% If the counterfactual at the beginning of this section holds, 
If we counteract the influence of deleting $z_-$ by adding $z_+$, then the model parameters after adding sample $z_+$ should be the same as the ones after deleting $z_-$, that is $ {\theta}^*_{\{z_+\}}= {\theta}^*_{\{z_-\}}$, put it into Eq. \ref{eq: parameters}, we can get:
\begin{equation}
    \frac{{1}}{{n}} {H}_{{\theta}^*}^{-1} \nabla_\theta {\ell}({z}_{-}, {\theta}^*) \approx-\frac{1}{n} {H}_{{\theta}^*}^{-1} \nabla_\theta {\ell}({z}_{+}, {\theta}^*).
    \label{eq: condition}
\end{equation}

To avoid the operation of calculating ${H}_{{\theta}^*}^{-1}$, we seek sufficient conditions that satisfy Eq.\ref{eq: condition}, and we have:
\begin{equation}
    \nabla_\theta {\ell}({z}_{-}, {\theta}^*)\approx -\nabla_\theta {\ell}({z}_{+}, {\theta}^*).
    \label{eq: sufficient_condition}
\end{equation}

Eq. \ref{eq: sufficient_condition} means that we can add a point with an opposite gradient to the deleted point to counteract its influence. Figuratively speaking, if we want to remove a point from linear regression, we can add another point on the opposite side to balance out the impact of the point to be deleted. Although Gradient Inversion Attacks \cite{zhu2019deep,geiping2020inverting} can provide us with the value of $z_+$ using gradient information, it is a time-consuming process. 
To this end, considering that Eq. \ref{eq: sufficient_condition} provides a gradient relationship between the forgotten sample and its ``opposite'' sample, it is natural to choose gradient descent for incremental learning, which allows us to cleverly avoid the challenge of generating ``opposite'' samples, making incremental learning easier. 
% \sout{It is undeniable that using more advanced incremental learning strategies is intriguing and meaningful, which will be considered part of our future research endeavors.}

Therefore, when we use gradient descent to incrementally learn the new sample $z_+$, then, according to Eq. \ref{eq: sufficient_condition}, 
it is equivalent to performing gradient ascent on $z_-$,
% the unlearning strategy seems to perform gradient ascent on $z_-$, 
that is,
$${\theta}^*_{unlearn}={\theta}^*+ \eta\cdot\nabla_\theta{\ell}({z}_{{-}}, \theta^*),$$ 
where $\eta$ is the learning rate. 
This incremental approximate learning method avoids retraining on $D-\{z_-\}$ and constructing $z_+$, saving time and effort. More importantly, this method does not require the calculation of the Hessian matrix and its inverse.
The gradient descent is only a means to achieve incremental learning, which allows us to cleverly avoid the challenge of generating "opposite" samples, making incremental learning easier. Undeniably, using more advanced incremental learning strategies is intriguing and meaningful, and it will be considered part of our future research endeavors.

% It is easy to see that this incremental approximate learning avoids retraining on $D-\{z_-\}$ \textcolor{blue}{and constructing $z_+$}. In addition, this approach saves time and effort without requiring the calculation of the Hessian matrix and its inverse, as rising once in the gradient of $z_-$ is sufficient.

Notably, this strategy can be easily generalized to batch deletions. If we want to unlearn a subset $D_f$ from the training set, the parameters change can be described by
\begin{equation}
    {\theta}^*_{unlearn}={\theta}^*+ \eta\cdot\sum_{{z}_i\in D_f}\nabla_\theta{\ell}({z}_{{i}}, \theta^*).
    \label{eq: approximation}
\end{equation}

\subsection{Gradient Correction}\label{sec: correction}
Only letting the parameters rise by the gradient of the unlearning point without a Hessian matrix as a weight constraint may cause the model to ``over-forget'' the unlearning point and ignore the gradients at the remaining points. 

Therefore, it is necessary to correct the update of the model on the gradient of the forgotten point to prevent this situation. Here, based on the ideal properties of machine unlearning introduced in Section \ref{sec: machine unlearning}, we suggest correcting the gradient in two directions: (1) according to the requirements of the comparable model utility, it should not damage the model performance at the remaining points, and (2) based on the removal guarantee, the model should keep forgetting the forgotten points.

This can be achieved by the idea of model catastrophic forgetting \cite{cata} subtly. Specifically, when a neural network trains on a new dataset, it will be more inclined to fit the new dataset, forget what it has previously learned, and cause the model to lose its previous capabilities. This inspires us to let the model strengthen the learning of the remaining training data, which can not only maintain the effect of the model on the remaining data but also consolidate the model's forgetting at the unlearned point $z_-$. Based on the above considerations, we get the corrected gradient as follows:
\begin{equation}
    \theta^*_{add}=\theta^*-\eta\cdot\sum_{{z}_i\in D_r} \nabla_\theta{\ell}({z}_{{i}}, \theta^*).
    \label{eq: correction}
\end{equation}
\subsection{Gradient Restriction}\label{sec: restriction}
Unlike traditional work that only focuses on the unlearning stage, we desire to improve the training quality of the model so that it has the potential to perform unlearning better and faster. 
This idea is also used in Unrolling SGD (USGD) \cite{unrolling}.
Outliers and abnormal points usually have large gradients on the model. If unlearn requests contain these points, simply letting the parameters rise by the gradient may destroy most of the information in the model. Since the forgetting point can be any data, we cannot ignore this phenomenon. So, we need to restrict the gradient of all points in the training dataset. 
Existing gradient-restricting techniques, such as Gradient Clip\cite{clip} and SignSGD\cite{signsgd}, mainly focus on correcting the gradient to prevent gradient explosion. However, they cannot guarantee that the real gradients of the sample are small enough. This leads to the fact that in the final trained model, the gradients of the unlearning samples may still be very large. We will further verify this in our experiments.

To this end, we propose a \textit{gradient restricted (GR) loss} in model training, which limits the gradients of the model to training samples not to be large, as shown below:
\begin{equation}
    {\ell}_{GR}({z}, \theta) = {\ell}({z}, \theta)+\alpha\cdot\|\nabla_\theta {\ell}({z}, \theta)\|_2.
    \label{eq: restriction}
\end{equation}
The new item can be seen as a regularization term, where $\alpha$ serves as the corresponding regularization coefficient.
By utilizing the empirical risk minimizer ${\theta}^* \stackrel{\text { def }}{=} \operatorname{argmin}_{\theta} \frac{1}{n} \sum_{i=1}^{n}
{\ell}({z}_{{i}}, \theta)$, we can conclude that $\nabla_{\theta^*}\sum_{i=1}^n\ell(z,\theta^*)=0$ \cite{guo2019certified, influence}. Therefore, when we regulate the first-order gradient $\nabla_\theta \ell(z,\theta)$, it results in $\nabla_\theta \sum_{i=1}^n\ell(z,\theta)$ tending towards 0, essentially leading to an optimal model with a more accurate direction.

The regularization term acts as a significant penalty on the gradients of parameters with high values on data points, ultimately favoring smaller, more uniform gradients. This property is highly beneficial as it encourages the network to use all points rather than just accommodating outliers and abnormal points. Therefore the model ends up with small gradients at all points, rather than large gradients at a few points.
Notably, although it is necessary to calculate the gradient and update it backward, the model is able to converge more quickly without significantly increasing the cost of model training. 
Our empirical results, as shown in Section~\ref{subsec:gr}, confirm these conclusions.

% \textbf{Differences from existing gradient-restricting methods}.   In contrast, the proposed Gradient Restriction module aims to ensure that the real gradient of each sample (without approximation by operations such as clip or sign) is not too large during the Model Training Phase. This will alleviate the damage to the unlearning model caused by the learning of abnormal samples in the Model Unlearning Phase, because the real gradient used is restricted in the previous Model Training Phase. We will empirically verify this in Appendix \ref{subsec:gr}.

\subsection{Overall Framework}
Through the above three modules, we achieve an approximation of the influence function and alleviate the time delay in calculating the Hessian matrix.

\textbf{Model Training Phase.} Based on Section \ref{sec: restriction}, we minimize the objective loss of Eq \ref{eq: restriction}. As we emphasized, gradient restriction helps the convergence of the model, and for this purpose, we use an early stopping mechanism. This makes the training delay of the overall model tolerable even though the computational complexity of a single round of training is higher than that of traditional training.

\textbf{Model Unlearning Phase.} Combining incremental approximation (Eq. \ref{eq: approximation}) and gradient correction (Eq. \ref{eq: correction}), we can get the unlearning strategy for model parameters updated as
\begin{equation}
    \theta^*_{unlearn}=\theta^*-\eta\cdot(\sum_{{z}_i\in D_r} \nabla_\theta {\ell}({z}_{{i}}, \theta^*)-\sum_{{z}_j\in D_f} \nabla_\theta{\ell}({z}_{{j}}, \theta^*)).
    \label{eq: unlearn}
\end{equation}
Specifically, when receiving an unlearning request, we let the model parameters increase on the gradient of $D_f$ and decrease on the one of $D_r$ according to the Eq. \ref{eq: unlearn}. Note that we only update model parameters once, i.e., we compute gradient on $\theta^*$ for the residual and forgotten sets, then update $\theta^*$ by these two gradients. Besides, our method is independent of the number of deletion points, which is very useful when receiving large batches of deletion requests. 

\textbf{Differences from gradient ascent unlearning methods.}
Although the existing strategies based on gradient ascent (descent) are common and intuitive, it is unclear how closely this heuristic strategy based on model learning relates to machine unlearning. Due to the wide application of unlearning in privacy protection, data security and other fields, this non-rigorous heuristic strategy may reduce users' trust in the unlearning model. On the contrary, inspired by the main theory of cognitive science that memorizing is often easier than forgetting \cite{wang2019more} and from the perspective of incremental learning, our work innovatively transforms “forgetting” into “memorizing" and establishes a bridge between incremental learning and machine unlearning through theoretical analysis (Section~\ref{subsection{Hessen matrix approximation}}). 
\subsection{Complexity Analysis}
Without specifying the model structure, assume that $t_1$ is the time for one forward propagation of the model, $k_1$ and $k_2$ are the maximum cost of computing an individual element of gradient and hessian matrix respectively, and $p$ is the number of model parameters. Upon receiving one unlearning request, the time complexity of the proposed strategy is $\mathcal{O}(nt_1+nk_1p)$. 

For the Hessian-based method, the time complexity of calculating the Hessian matrix is $\mathcal{O}(nt_1+nk_2p^2)$. The time complexity of directly calculating the inverse of the Hessian matrix is $\mathcal{O}(p^3)$, so the total time complexity of the Hessian-based method is $\mathcal{O}(nt_1+nk_2p^2+p^3)$, which is much larger than the proposed algorithm. Even if the inverse of the hessian matrix can be approximated through numerical optimization algorithms such as L-BFGS or Lissa and the complexity can be reduced to $\mathcal{O}(nt_1+nk_2p^2+tp)$, where $t$ is the number of optimizations, it is still much larger than the proposed IAU.
\section{Experiment}
\subsection{Experimental Setup}
\subsubsection{Datasets} 
% We conducted our experiments on CIFAR10 \cite{krizhevsky2009learning} and SVHN \cite{netzer2011reading}. These datasets are widely used for evaluating the performance of deep neural networks \cite{badtea,unrolling,zhang2022prompt,sisa}. 
% We also conducted experiments on tabular dataset Purchase100 in Appendix \ref{app:ta} and complex dataset CIFAR100 \cite{krizhevsky2009learning} in Appendix \ref{app:100}. More details of these datasets can be found in Appendix \ref{app:data}.
We conducted experiments on CIFAR10 \cite{krizhevsky2009learning} and SVHN \cite{netzer2011reading}, widely used datasets for evaluating the performance of deep neural networks \cite{badtea,unrolling,zhang2022prompt,sisa}. Additionally, we also explored our IAU algorithm on tabular dataset Purchase100 and more complex dataset CIFAR100 \cite{krizhevsky2009learning}, which allowed us to assess the versatility of our approach across diverse data formats and levels of complexity. 
% CIFAR10 consists of 60,000 32x32 color images categorized into 10 classes, providing a fundamental benchmark for image classification. CIFAR100 extends this challenge with 100 classes, each containing 600 images. The SVHN dataset features over 600,000 images of digits (0-9) extracted from Google Street View, making it valuable for real-world digit recognition tasks. Purchase100, a dataset comprising 197,324 samples, includes 600 binary indicators representing individual product purchases and enables the prediction of shopping preferences.

\subsubsection{Models}
We conduct comparisons of unlearning methods using two widely-used models \cite{zhang2022prompt,amne,unrolling}: LeNet5 \cite{lecun1998gradient} and ResNet18 \cite{he2016deep}. LeNet5 consists of two convolutional layers, followed by max-pooling and then two fully connected layers. ResNet-18 consists of 18 weight layers, including a 7x7 convolutional layer, four residual blocks, and fully connected layers. Additionally, we conducted experiments on MLP \cite{mlp} in Section \ref{app:ta} and VGG19 \cite{simonyan2015very} in Section \ref{app:100} to further assess the versatility of our unlearning approaches across diverse model architectures.
% We conduct comparisons of unlearning methods using two widely-used models \cite{zhang2022prompt,amne,unrolling}: LeNet5 \cite{lecun1998gradient} and ResNet18 \cite{he2016deep}. LeNet5 consists of two convolutional layers, followed by max-pooling and then two fully connected layers. ResNet-18 consists of 18 weight layers, including a 7x7 convolutional layer, four residual blocks, and fully connected layers. We also conducted experiments on MLP \cite{mlp} in Appendix \ref{app:ta} and VGG19 \cite{simonyan2015very} in Appendix \ref{app:100}.

\subsubsection{Evaluation Metrics}
The primary goal of approximate unlearning is to ensure that the distribution of final activation results from the unlearned model closely resembles that of the retrained model, making them nearly indistinguishable. 
Consequently, we regard the retrained model as the gold standard for evaluating unlearning methods based on the following three criteria, which collectively offer a comprehensive evaluation of the unlearning model's similarity to the retrained model and the efficiency of the unlearning method.
The following three metrics, namely Model Utility (MU), Unlearning Time (Time), and Unlearning Efficacy (UE), are specifically tailored to be consistent with the three ideal unlearning properties (Removal Guarantee, Unlearning Efficiency, and Comparable Model Utility discussed in Section \ref{sec: machine unlearning}) respectively. In addition, their combined consideration (i.e., Avg Rank) will provide a comprehensive assessment framework for evaluating the effectiveness of an unlearning algorithm.
\begin{itemize}[leftmargin = 10 pt]
    \item \textbf{Model Utility (MU).} For an effective unlearning algorithm, it is imperative that the MU closely approximates that of the retrained model. This alignment can be quantitatively assessed by measuring the gap between the accuracy of the test dataset achieved by the unlearning model and that of the golden model.
    % between the accuracy \textcolor{blue}{on test dataset} of the retrained model and the accuracy of the unlearning model. 
    A low MU value indicates minimal deviation from the retrained model's performance.
    \item \textbf{Unlearning Time (Time).} The unlearning mechanism should be time-efficient compared to retraining. We record the time consumed by the unlearning algorithm, serving for the quantitative evaluation of Unlearning Efficiency. Greater efficiency is achieved with shorter Unlearning Time.
    % We record the unlearning time to reflect the unlearning efficiency across different unlearning algorithms. 
    \item \textbf{Unlearning Efficacy (UE).} From the attacker's perspective, the unlearning model should closely resemble the retrained model. 
    UE is to quantify the degree of proximity between the unlearning model and the golden model, as perceived by a potential attacker. It is quantified by calculating the gap in the attack success rate concerning the erased dataset between the unlearning model and the golden model, typically through the use of a Membership Inference Attack (MIA) \cite{yeom2018privacy}. A lower UE value signifies a heightened degree of resemblance between the unlearning model and the golden model, as perceived by the potential attacker.
    % Unlearning Efficacy (UE) can be quantified by calculating the gap between the attack success rate on the unlearning model and the retrained model, typically through the use of a Membership Inference Attack (MIA) \cite{yeom2018privacy}. This measurement assesses the ability of the unlearning process to maintain the model's privacy and security characteristics.
    \item \textbf{Average Rank(Avg Rank).} When evaluating the unlearning algorithm, three essential criteria are considered: MU, Time, and UE. Each unlearning algorithm aims to achieve a balance or trade-off between these dimensions. To thoroughly evaluate the algorithm's performance, we use ``Avg Rank'' as a composite metric that reflects the average ranking across all three dimensions. A lower rank indicates a better trade-off achieved by the unlearning algorithm, and an ideal unlearning algorithm would have a rank of 0.
\end{itemize}

\begin{table*}[t]
\centering
\renewcommand{\arraystretch}{1.2}
\caption{Comparison of Model Utility(MU), unlearning time, and Unlearning Efficacy(UE) with baselines for 5\% points unlearned randomly from the original training points in two datasets and two ML models. The optimal approach for these three indicators is to minimize their values. The optimal outcomes for each backbone on each dataset are represented in bold typeface, while the second-best outcomes are indicated with an underline.}
\resizebox{\linewidth}{!}{
\begin{tabular}{c|c|cccc|cccc}
\hline
\multirow{2}{*}{Backbone} & \multirow{2}{*}{Strategy} & \multicolumn{4}{c|}{CIFAR10}   & \multicolumn{4}{c}{SVHN}      \\
\cline{3-10}
                          &                     & MU$\downarrow$  & Time(second)$\downarrow$  & UE$\downarrow$     & Avg rank$\downarrow$     & MU$\downarrow$   & Time(second)$\downarrow$  & UE$\downarrow$     & Avg rank$\downarrow$    \\ \hline
\multirow{6}{*}{LeNet5}   & Retrain             & 0    & 414          & 0     &              & 0    & 822          & 0     &              \\
                          & USGD       & 0.80 & 33           & 2.27  & {\uline{1.7}}    & 5.16 & 20           & 6.17  & 3.3          \\
                          & Bad Teaching        & 1.38 & 23           & 6.11  & 2.7          & 3.10 & 14           & 2.17  & {\uline{1.3}}    \\
                          & Amnesiac Unlearning       & 0.51 & 33 & 3.21 & \textbf{1} & 0.04 & 26 & 1.67 & \textbf{1} \\
                          & Fisher              & 0.61 & 1294         & 8.08  & 3            & 4.53 & 1926         & 4.75  & 3.3          \\
                          & IAU(Ours)           & 1.31 & 13           & 5.21  & {\uline{1.7}}    & 0.09 & 10           & 2.46  & \textbf{1}   \\ \hline
\multirow{6}{*}{ResNet18} & Retrain             & 0    & 424          & 0     &              & 0    & 575          & 0     &              \\
                          & USGD       & 1.52 & 27           & 13.98 & {\uline{1.7}}    & 0.07 & 43           & 4.99  & 2            \\
                          & Bad Teaching        & 0.98 & 20           & 64.93 & 2            & 0.01 & 20           & 4.26  & \textbf{0.7} \\
                          & Amnesiac Unlearning & 5.13 & 39           & 64.80 & 3.3          & 1.64 & 50           & 20.95 & 3.7          \\
                          & Fisher              & 1.51 & 3078         & 24.74 & 2.7          & 0.02 & 4503         & 5.04  & 2.7          \\
                          & IAU(Ours)           & 0.42 & 19           & 20.10 & \textbf{0.3} & 0.74 & 12           & 3.10  & {\uline 1}      \\ \hline
\end{tabular}
}
\label{maintable}
\end{table*}

\subsubsection{Baselines}
We implement the following baseline unlearning methods for comparisons:
\begin{itemize}[leftmargin = 10 pt]
    \item \textbf{Retrain}. We train the model from scratch with the remaining data as the retrained model. Thus, the retrained model is the optimal unlearned model and is seen as the gold model.
    \item \textbf{Unrolling SGD (USGD)} \cite{unrolling}. USGD uses the standard deviation (SD) loss in the training framework for pre-training epochs, then trains additional epochs on the subset of the training set and records gradient. In the unlearning phase, it resumes the gradient decreased by unlearning points.
    \item \textbf{Amnesiac Unlearning} \cite{amne}. Amnesiac Unlearning removes unlearning examples and inserts a small number of copies of them with randomly selected incorrect labels. Then, it fine-tunes the model with those random labels on forgotten samples.
    \item \textbf{Bad Teaching} \cite{badtea}. Bad teaching explores the utility of competent and incompetent teachers in a student-teacher framework to induce forgetfulness. The knowledge from the competent and incompetent teachers is selectively transferred to the student to obtain a model that does not contain any information about the forgotten data.
    \item \textbf{Fisher} \cite{fisher}. Fisher locates the influence of unlearning points by using the Fisher Information Matrix as a Hessian approximation. Then, it scrubs the influence of the unlearning points on model parameters.
\end{itemize}
% We compare IAU with \textbf{Unrolling SGD (USGD)} \cite{unrolling}, \textbf{Amnesiac Unlearning} \cite{amne}, \textbf{Bad Teaching} \cite{badtea}, and \textbf{Fisher} \cite{fisher}. Their details can be found in Appendix \ref{sec: appendix_baselines}

\subsubsection{Membership Inference Attack Details}
\label{mem}
We adopt the Membership Inference Attack(MIA) proposed in \cite{shokri2017membership}. Specifically, In our scenario, the adversary only has black-box access to the target model, meaning that the adversary can submit a data point to the target model and subsequently obtain the probabilistic output. Furthermore, the attacker is privy to the architecture of the victim model and has access to data distributions identical to those used during the victim model's training process. Consequently, the attacker leverages this knowledge to construct multiple shadow models that mirror the behavior of the victim model.
In particular, we have trained three shadow models, each of which shares the same structure as the target model. The dataset employed for training the shadow models is drawn from the same distribution as the training data used for the victim model. The attack model is designed as a fully connected network with two hidden layers featuring widths of 256 and 128, respectively. ReLU activation functions, dropout layers with a rate of 0.5, and a sigmoid output layer are incorporated into this architecture. The attack model is applier to the model on the forget dataset  $D_f$ to calculating UE.

\subsubsection{Implementation Details}
We conduct LeNet experiments on a single Nvidia RTX 3090 GPU server with Intel Xeon CPUs  ResNet18 and VGG19 experiments on two Nvidia A100 GPU servers with Intel Xeon CPUs and MLP experiments on one Nvidia RTX 2080 Ti GPU servers with Intel Xeon CPUs. We implemented and conducted training using the PyTorch deep learning framework, version 2.0.1. All experiments have been conducted 10 times, and the reported results represent the average values across these repetitions. To mitigate overfitting, we employ an early stopping mechanism. Specifically, if there is no improvement in validation set accuracy for 10 epochs, we terminate the model's training. 
% Our code is available at https://github.com/Lolo1222/IAU .
\subsection{Comparison with Baselines}
% Table \ref{maintable} shows the results of comparison with baselines for 5\% points unlearned randomly from the original training points in two datasets and two ML models. 
TABLE \ref{maintable} shows the results of comparison with baselines with 5\% of training points randomly unlearned. From the table, we have the following important findings.
% The comparison results conducted on two machine learning models, namely LeNet5 and ResNet18, across two datasets - CIFAR10 and SVHN, with 5\% of training points randomly unlearned, are displayed in Table \ref{maintable}. These results have revealed following significant findings. Additionally, we have included the experimental results of removing outliers in the Appendix \ref{app:out}.
% From them, we have the following important findings.
\begin{itemize}[leftmargin = 10 pt]
    \item[$\bullet$]  \textit{IAU consistently outperforms the four state-of-the-art unlearning baselines in terms of removal guarantee, unlearning efficiency, and comparable model utility.
    } The results clearly demonstrate that our method outperformed the baselines in the two experiments LeNet5 on CIFAR10 and ResNet18 on SVHN, achieving the top ranking. In the remaining two experiments, our method secured the second position, still showing a strong performance that surpassed the other baselines. This consistent pattern of success across the experiments underlines the superior performance of our method, highlighting its effectiveness and robustness in outperforming existing approaches for different datasets and models.
    % , securing the top rank in two experiments and the second position in the other two.
    \item[$\bullet$]  \textit{IAU demonstrates exceptional performance in terms of time efficiency, outperforming all baseline methods in every experiment, and it can strike a superior balance between unlearning efficiency and model utility.}
    For the performance of unlearning efficiency, IAU efficiently minimizes unlearning time. For example, on the LeNet5 model, IAU outperforms the second fastest method by around 28.6\%$\sim$43.5\%. This accomplishment is particularly noteworthy in situations where deletion requests are frequent. 
   In terms of model utility, the IAU algorithm yields results within a margin of 2 units from the retraining process, a feat unattained by any of the other baseline methods. 
    These findings unequivocally establish IAU's ability to balance performance and efficiency, consistently surpassing established baseline techniques. 
    \item[$\bullet$]\textit{IAU consistently achieves comparable performance to Fisher in terms of unlearning effectiveness while significantly reducing the unlearning time by several orders of magnitude.}  Both approaches are dedicated to alleviating the influence of unlearning data points on model parameters through the utilization of influence functions, thus presenting a commonality in their efficacy for unlearning. Nonetheless, the pivotal divergence emerges with Fisher's approach, which entails the approximation of the Hessian matrix by employing the Fisher Information Matrix and subsequently inverting it. In contrast, our proposed IAU methodology adeptly circumvents this computational step. Hence, IAU exhibits superior time efficiency while retaining a good forgetting effect compared to Fisher.
\end{itemize}
\subsection{Unlearning Efficacy}

\begin{figure*}[htbp]
    \centering
    \begin{minipage}{0.11\linewidth}
    \centering
    \includegraphics[width=\linewidth]{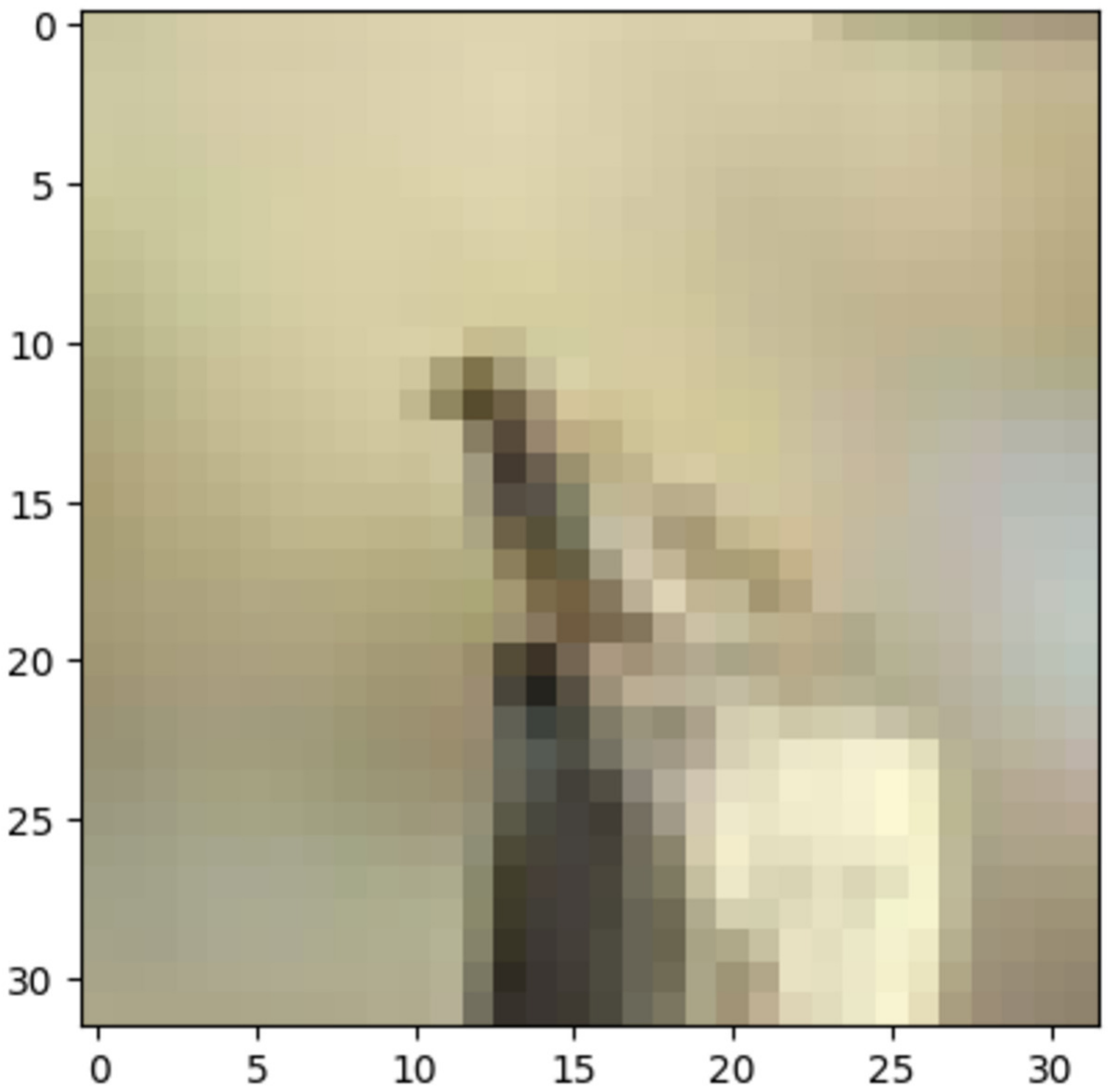}
    \vspace{-1.0cm}
    \caption*{Unlearned}
    \end{minipage}
    \begin{minipage}{0.11\linewidth}
    \centering
    \includegraphics[width=\linewidth]{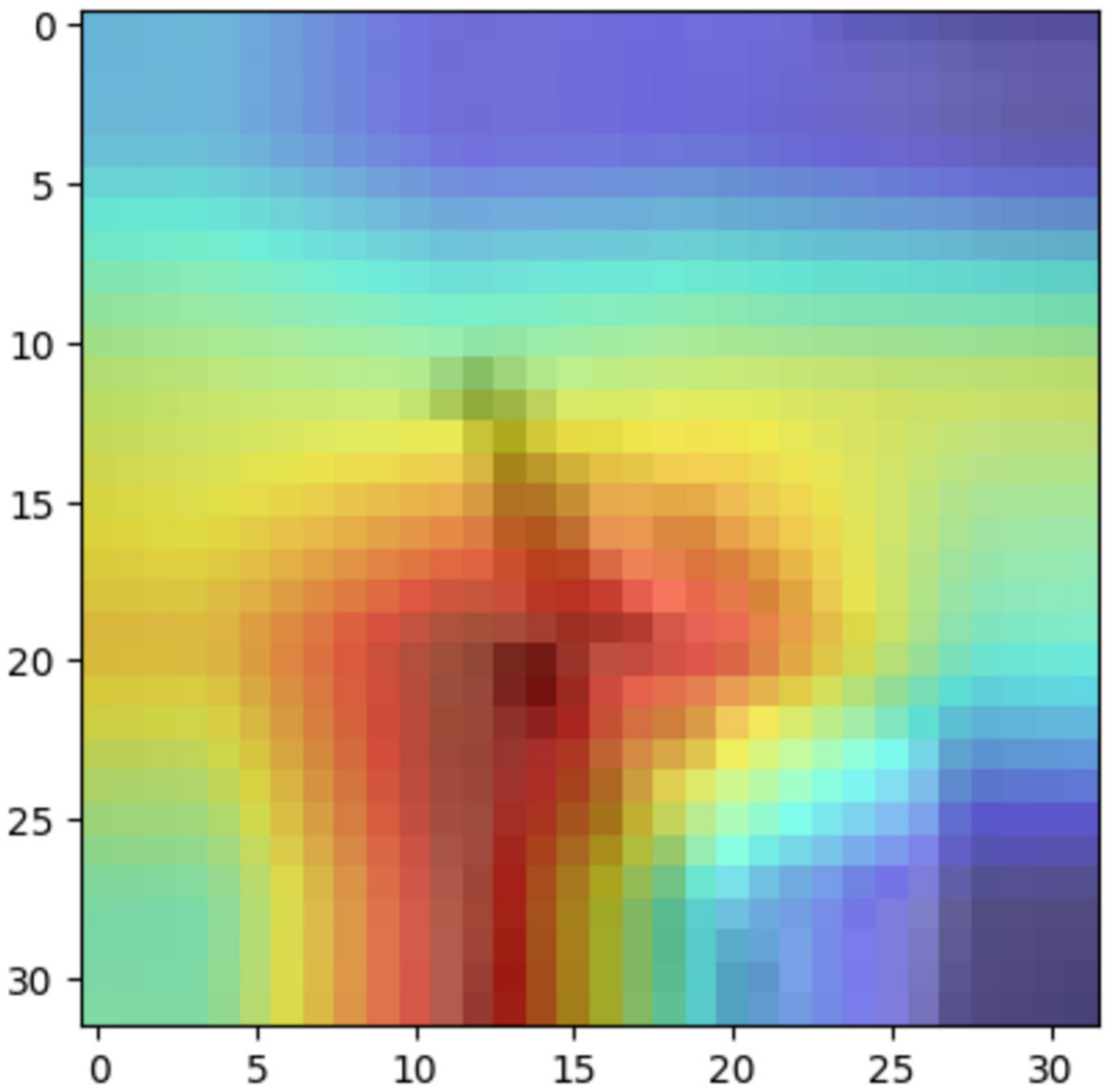}
    \vspace{-1.0cm}
    \caption*{Before}
    \end{minipage}
    \begin{minipage}{0.11\linewidth}
    \centering
    \includegraphics[width=\linewidth]{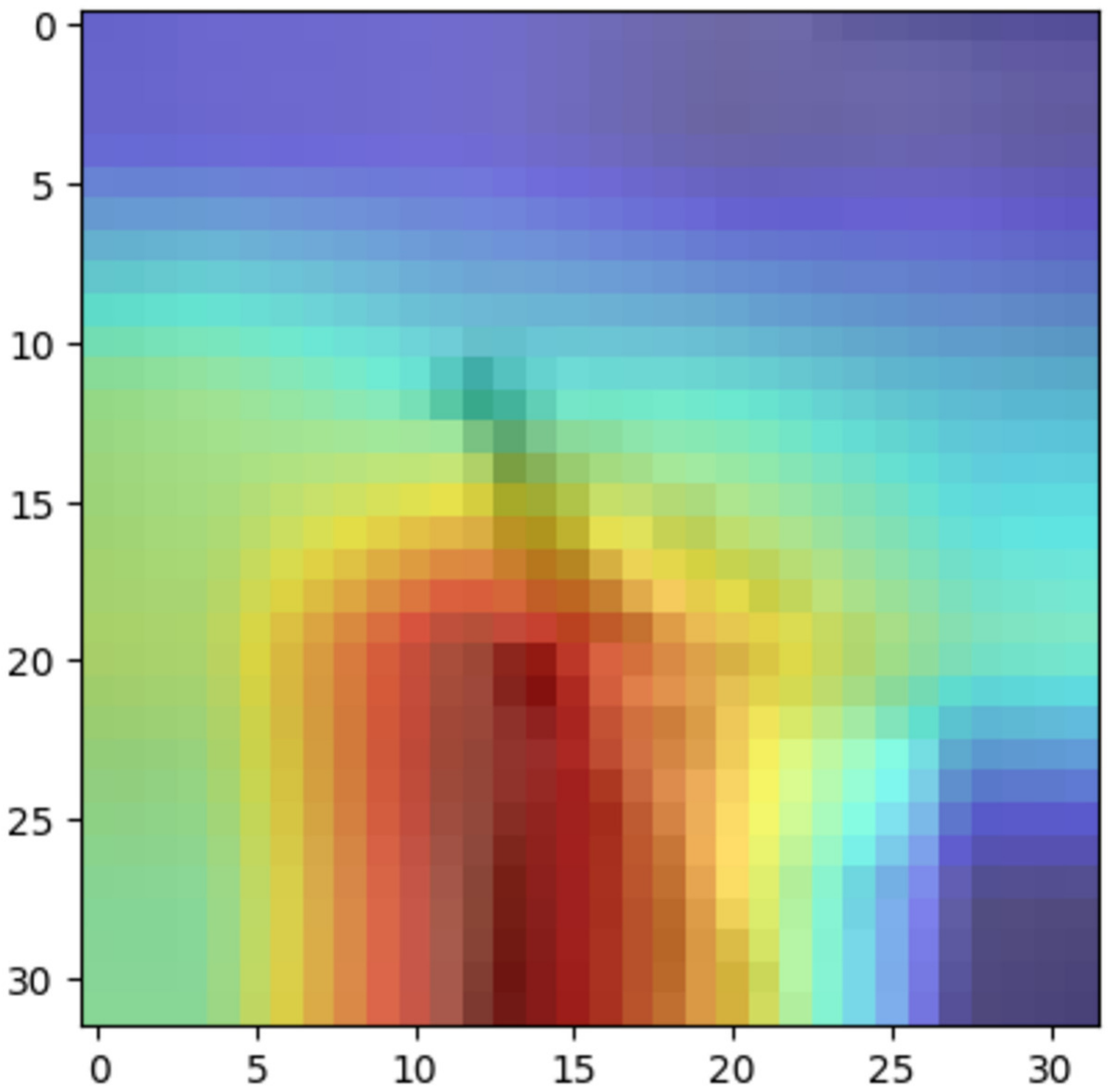}
    \vspace{-1.0cm}
    \caption*{Retrain}
    \end{minipage}
    \begin{minipage}{0.11\linewidth}
    \centering
    \includegraphics[width=\linewidth]{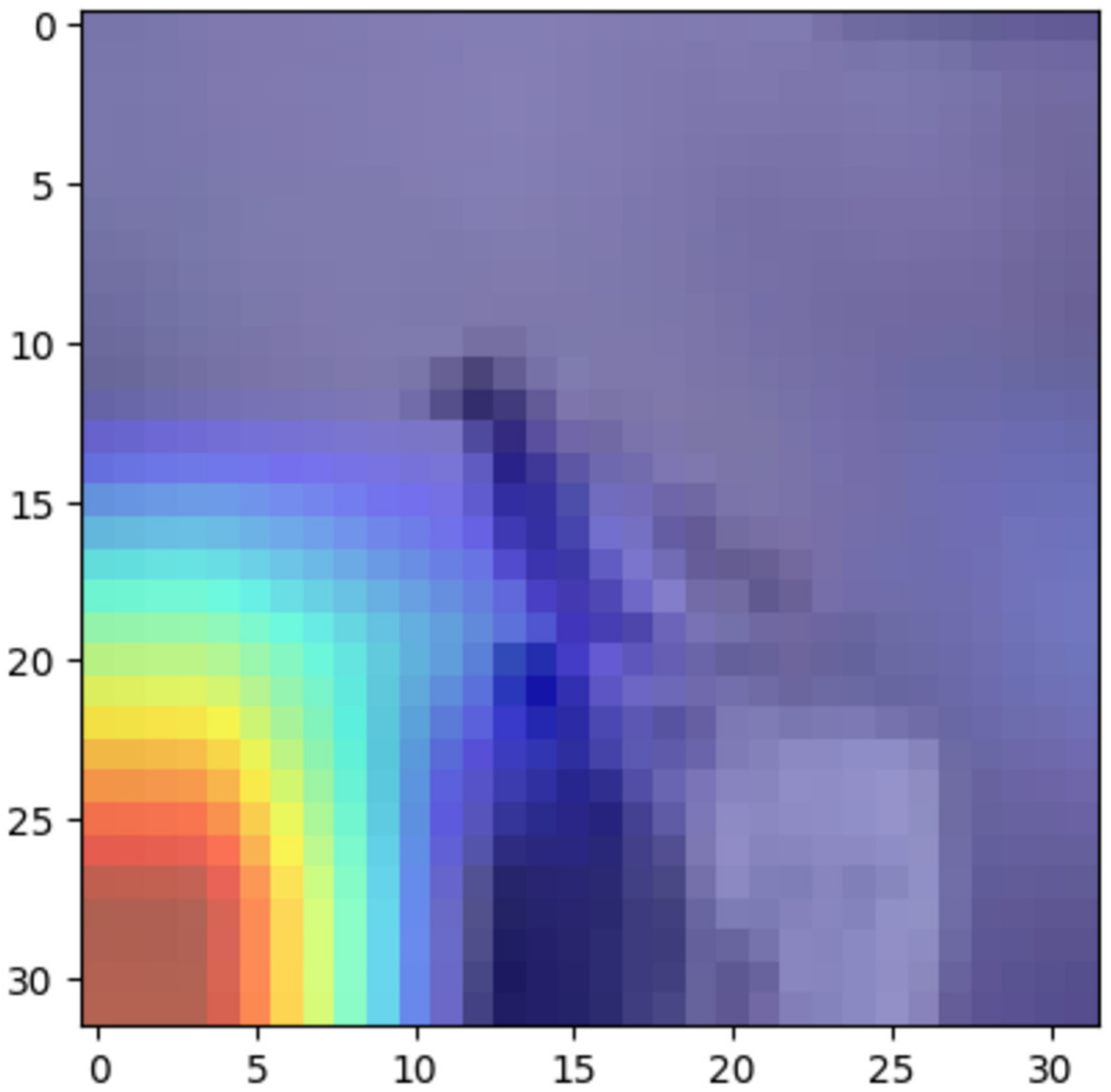}
    \vspace{-1.0cm}
    \caption*{{USGD}}
    \end{minipage}
    \begin{minipage}{0.11\linewidth}
    \centering
    \includegraphics[width=\linewidth]{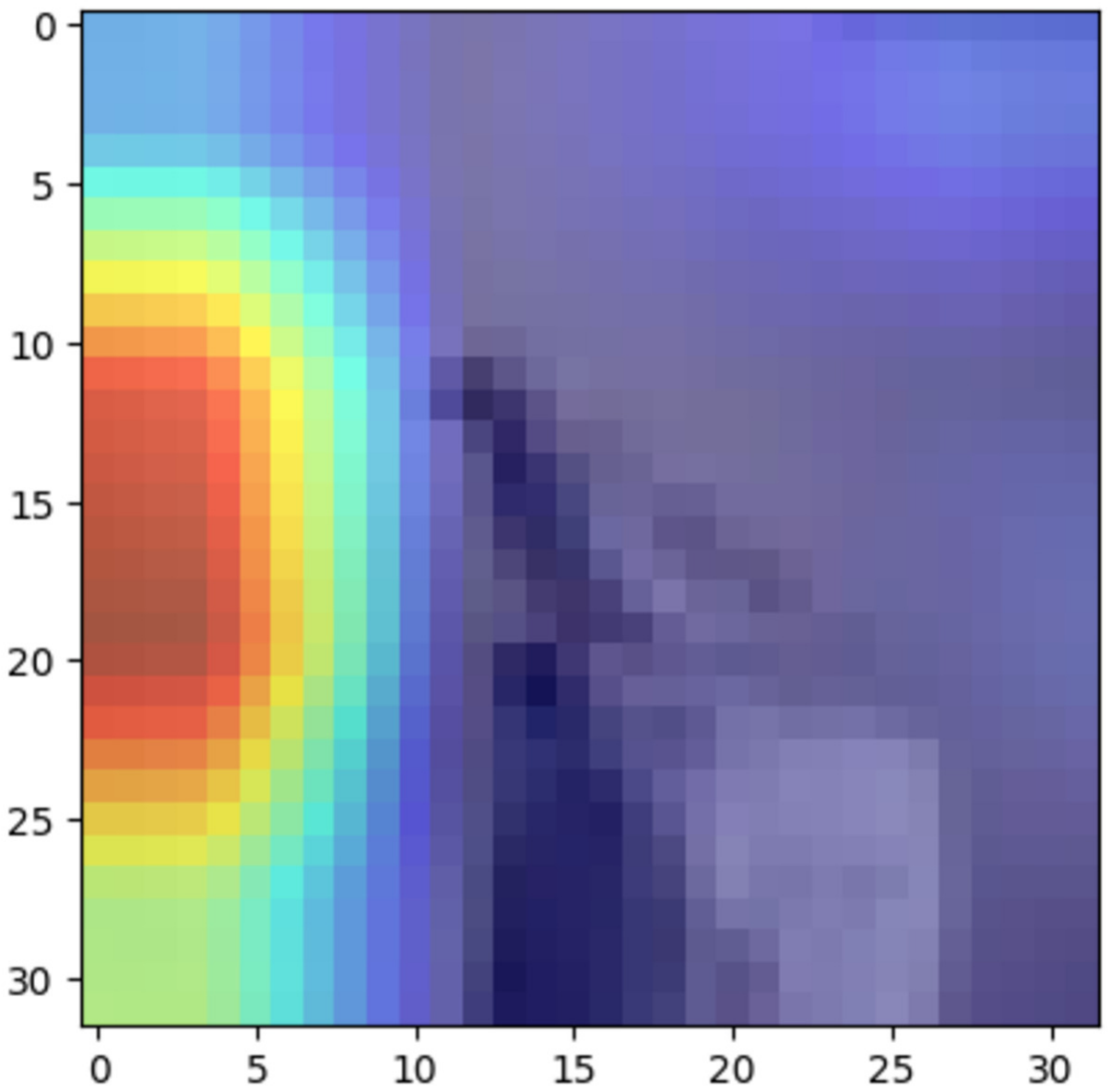}
    \vspace{-1.0cm}
    \caption*{Badteaching}
    \end{minipage}
    \begin{minipage}{0.11  \linewidth}
    \centering
    \includegraphics[width=\linewidth]{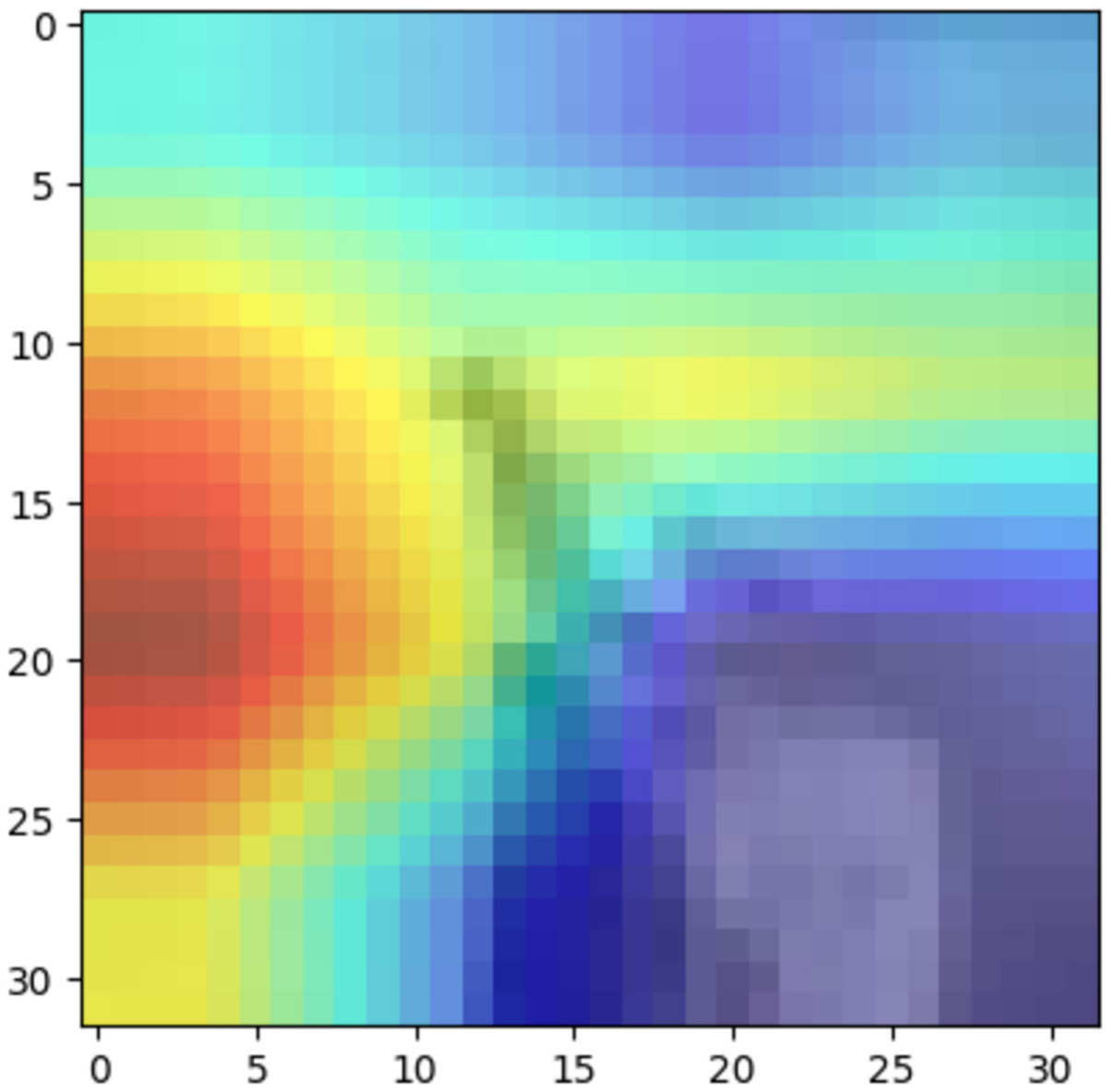}
    \vspace{-1.0cm}
    \caption*{Amnesiac}
    \end{minipage}
    \begin{minipage}{0.11  \linewidth}
    \centering
    \includegraphics[width=\linewidth]{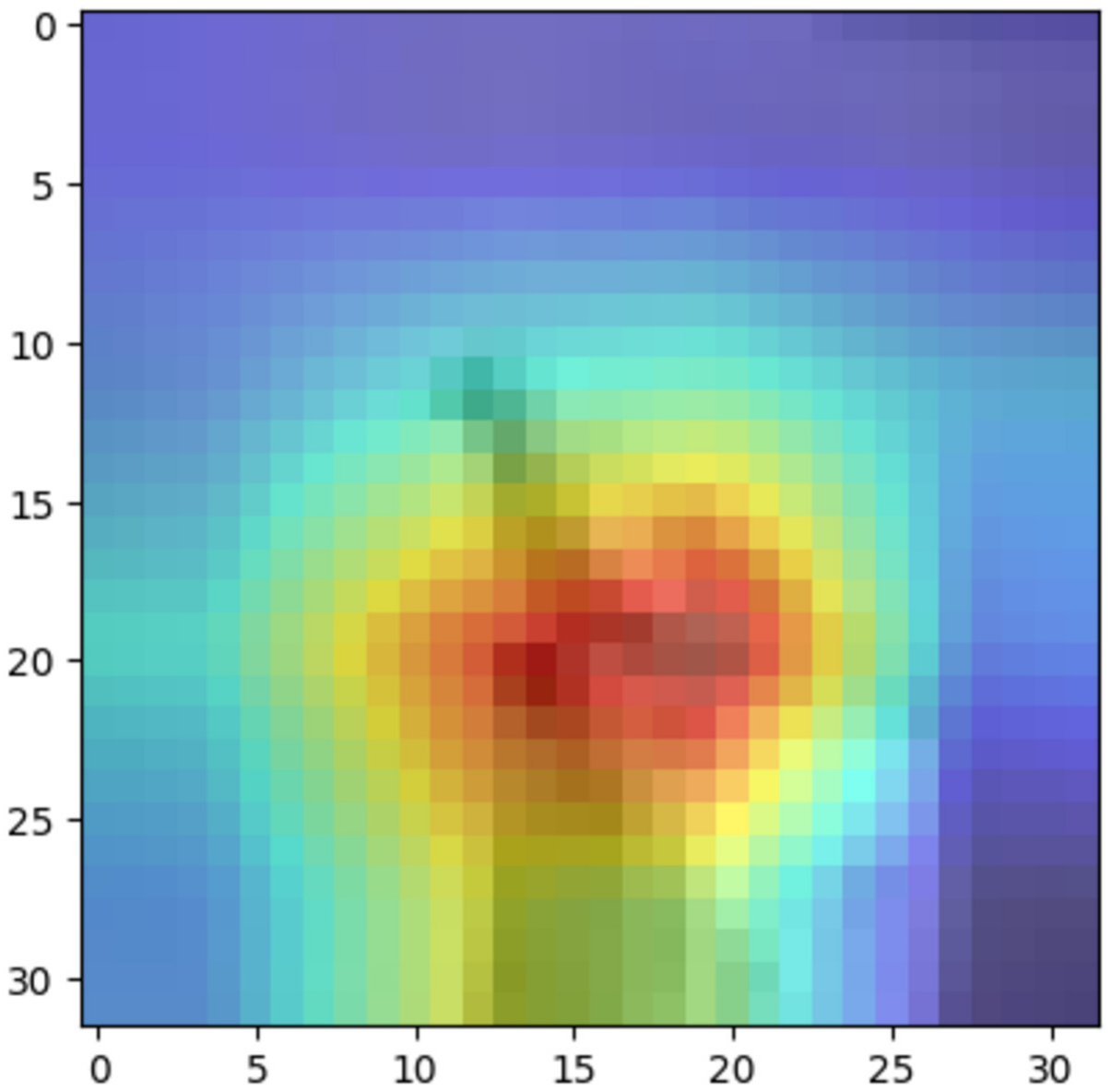}
    \vspace{-1.0cm}
    \caption*{Fisher}
    \end{minipage}
    \begin{minipage}{0.11  \linewidth}
    \centering
    \includegraphics[width=\linewidth]{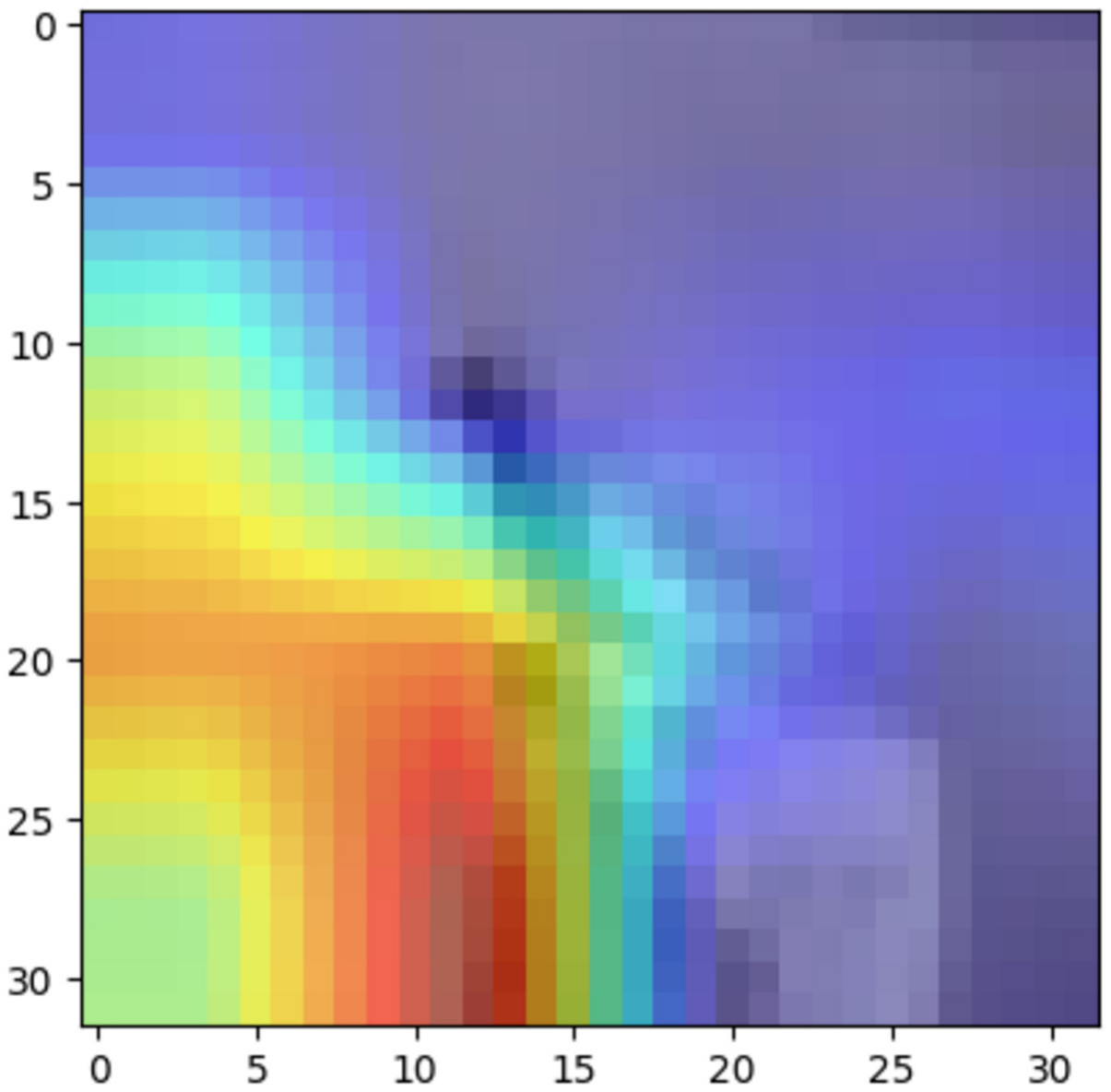}
    \vspace{-1.0cm}
    \caption*{IAU(Ours)}
    \end{minipage}
    
    \begin{minipage}{0.11  \linewidth}
    \centering
    \includegraphics[width=\linewidth]{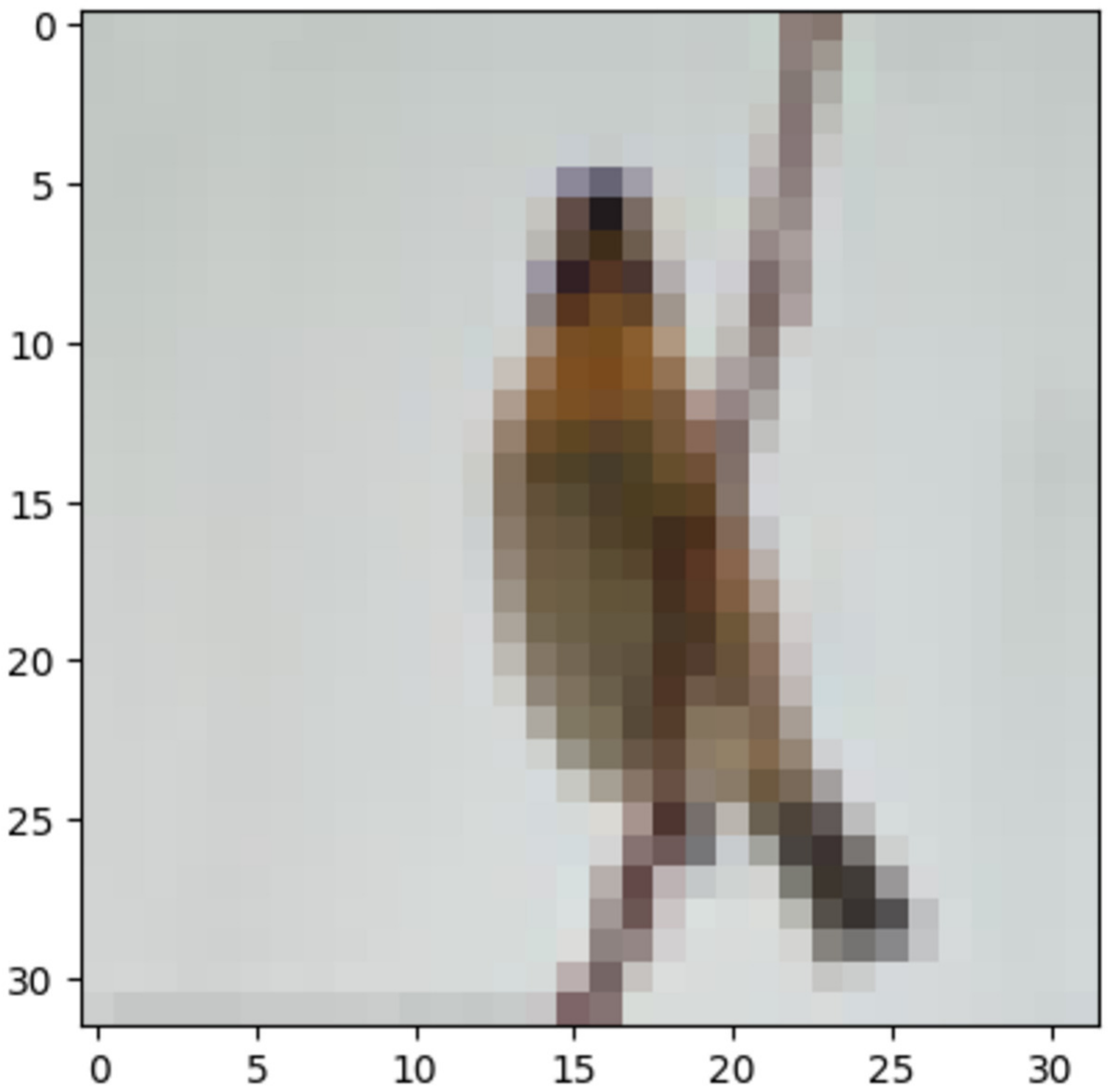}
    \vspace{-1.0cm}
    \caption*{Retained}
    \end{minipage}
    \begin{minipage}{0.11  \linewidth}
    \centering
    \includegraphics[width=\linewidth]{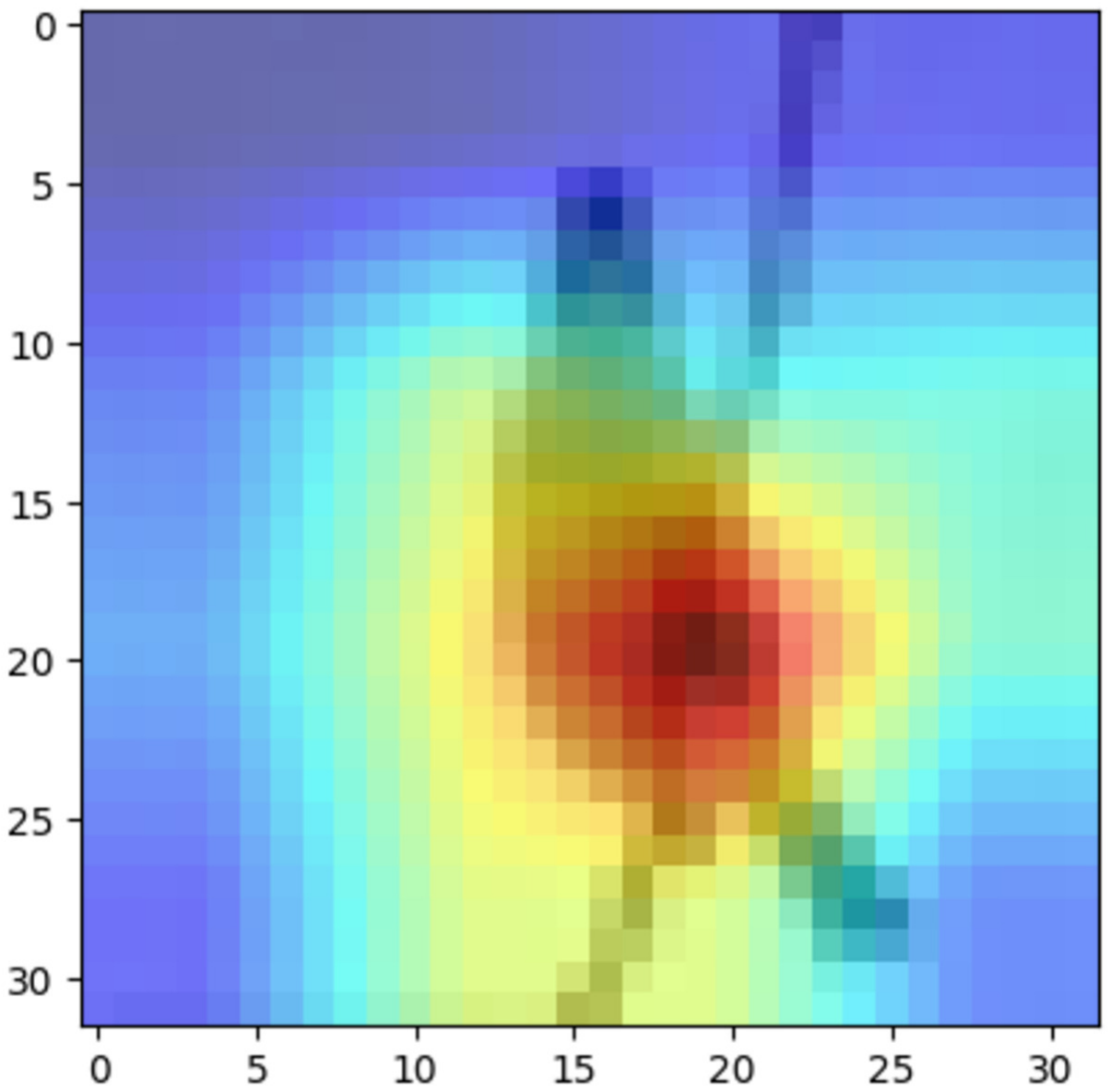}
    \vspace{-1.0cm}
    \caption*{Before}
    \end{minipage}
    \begin{minipage}{0.11  \linewidth}
    \centering
    \includegraphics[width=\linewidth]{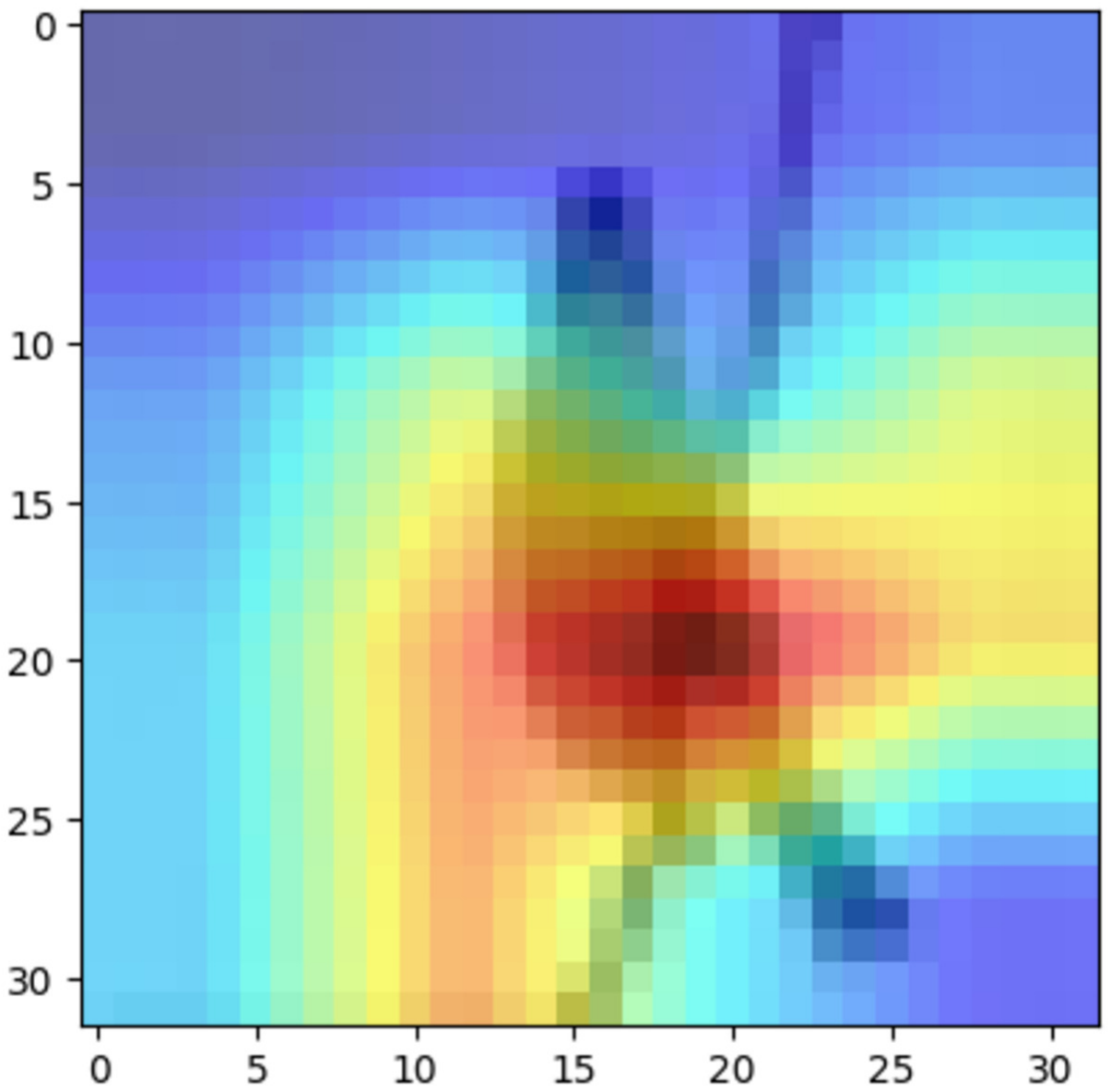}
    \vspace{-1.0cm}
    \caption*{Retrain}
    \end{minipage}
    \begin{minipage}{0.11  \linewidth}
    \centering
    \includegraphics[width=\linewidth]{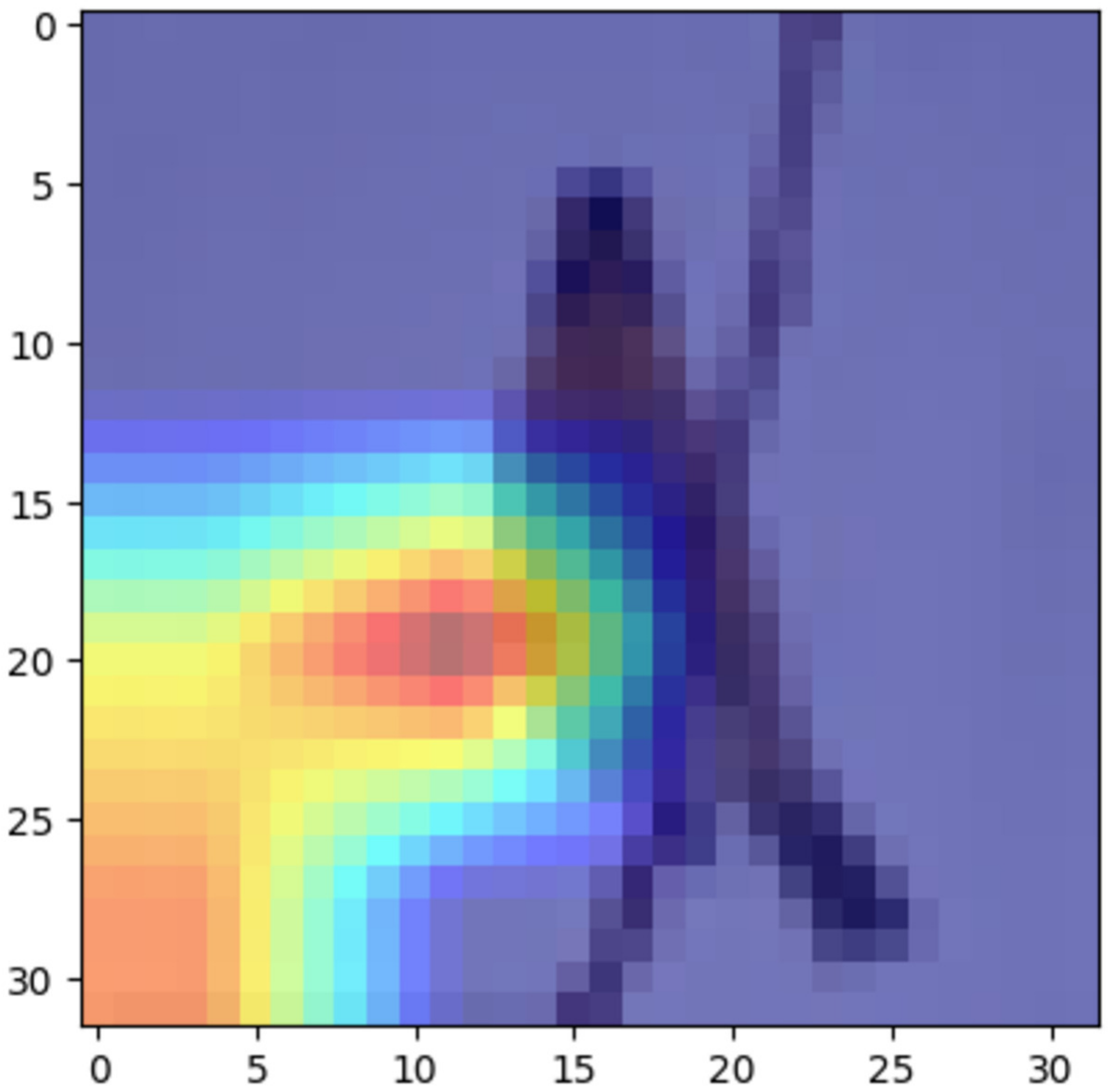}
    \vspace{-1.0cm}
    \caption*{{USGD}}
    \end{minipage}
    \begin{minipage}{0.11  \linewidth}
    \centering
    \includegraphics[width=\linewidth]{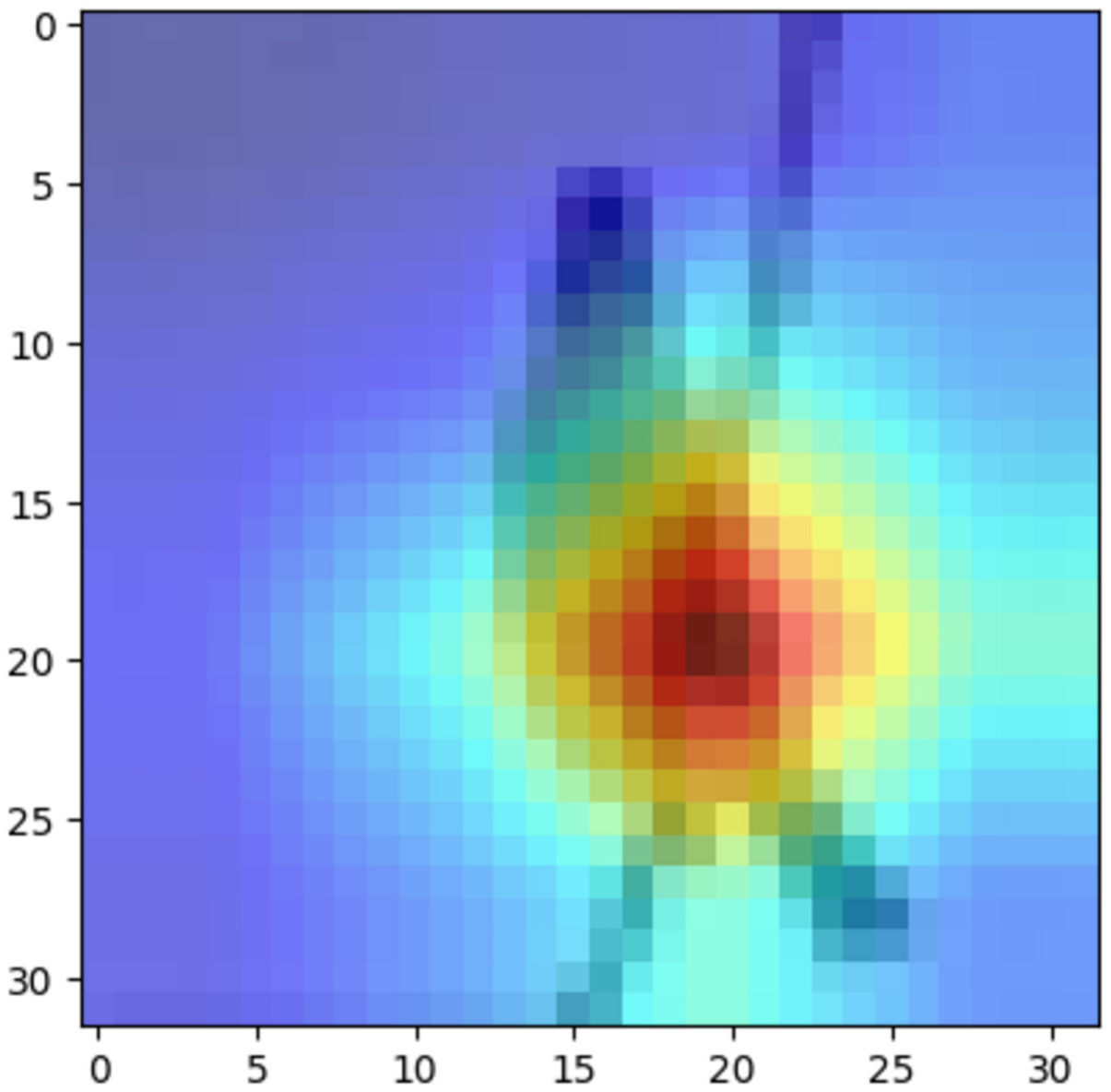}
    \vspace{-1.0cm}
    \caption*{Badteaching}
    \end{minipage}
    \begin{minipage}{0.11  \linewidth}
    \centering
    \includegraphics[width=\linewidth]{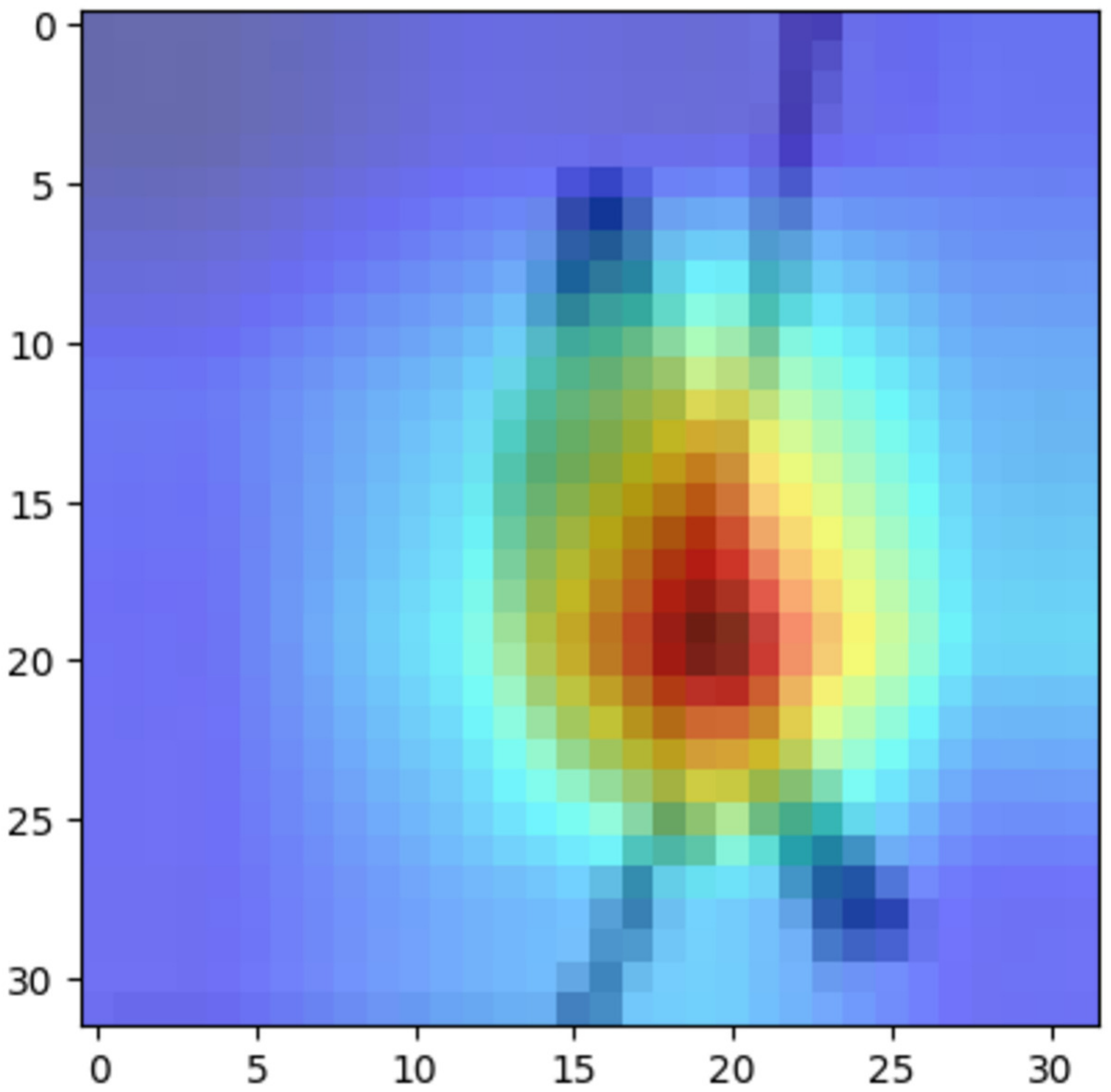}
    \vspace{-1.0cm}
    \caption*{Amnesiac}
    \end{minipage}
    \begin{minipage}{0.11  \linewidth}
    \centering
    \includegraphics[width=\linewidth]{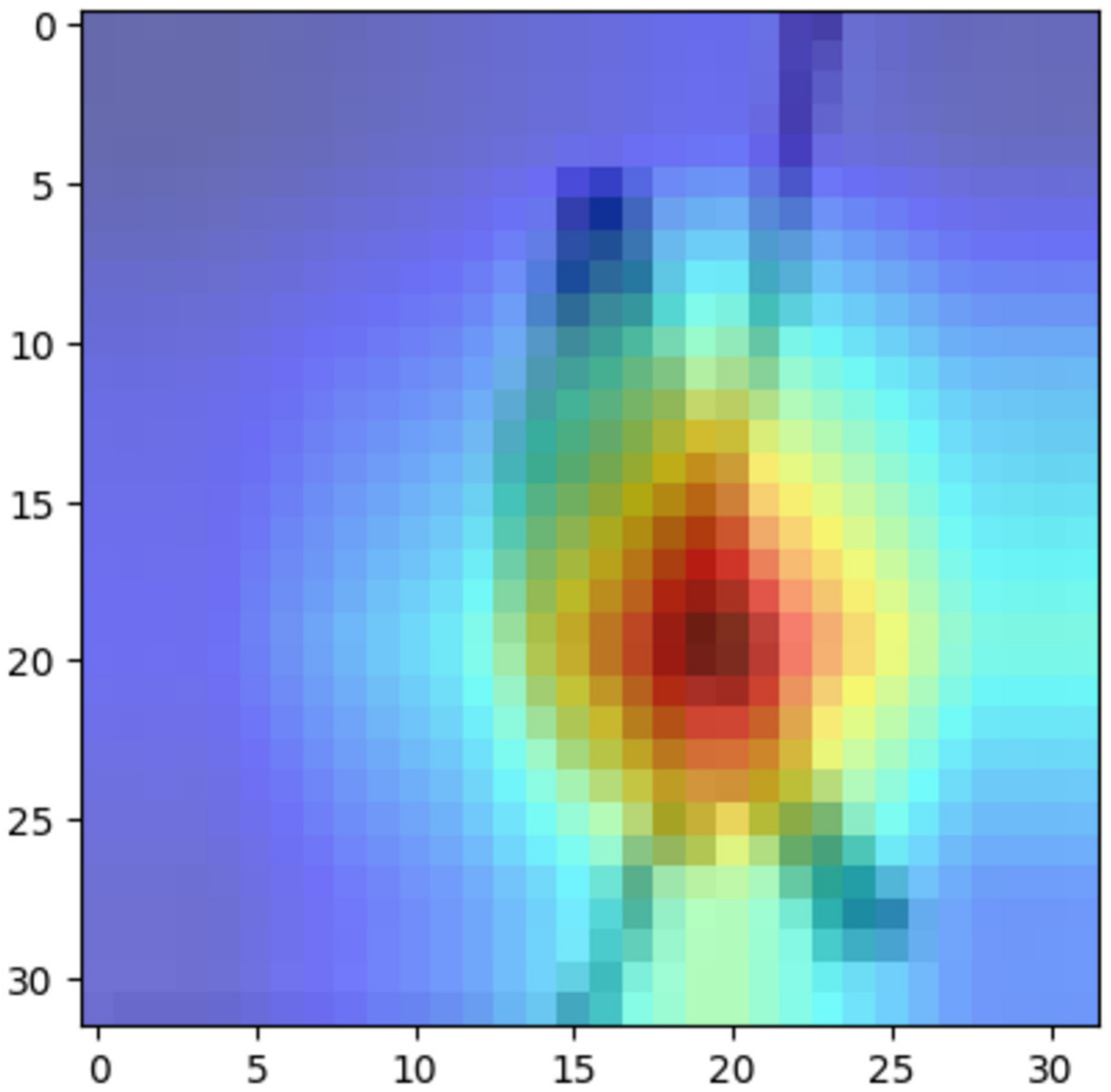}
    \vspace{-1.0cm}
    \caption*{Fisher}
    \end{minipage}
    \begin{minipage}{0.11  \linewidth}
    \centering
    \includegraphics[width=\linewidth]{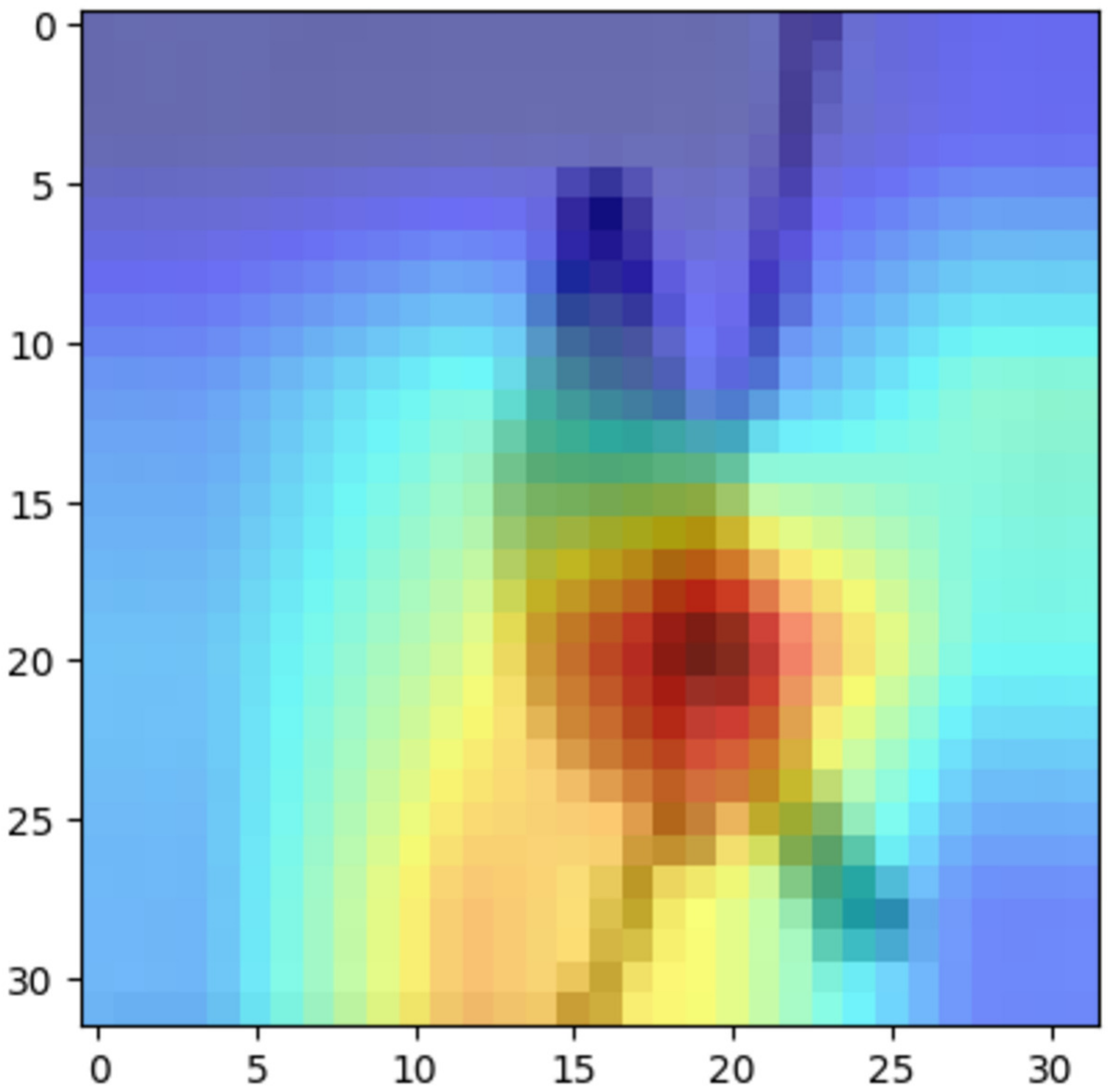}
    \vspace{-1.0cm}
    \caption*{IAU(Ours)}
    \end{minipage}
    
    \caption{These are activation maps that demonstrate the impact of scrubbing on a model. The top row displays the scrubbed image, while the bottom row displays a non-scrubbed image. Each row shows the original image, the activation map before unlearning, and the activation map after the unlearning method. The method names are noted below the activation map obtained using each method.}
    \label{gradcam}
\end{figure*}
We conducted an evaluation of ML unlearning methods on deep neural networks that were trained for image classification. Our experiments were performed on the CIFAR10 dataset in the ResNet18 model
% that was trained on the entire set of training images. 
To observe the unlearning performance, we randomly selected one training image and applied unlearning to it. The results are as shown in Fig.\ref{gradcam}. 

It is observable that the activation map of the unlearned image, as obtained through retraining, exhibits a slight downward shift rather than a significant alteration. This outcome is in line with expectations since the unlearning point is not orthogonal to the remaining data points.

The activation maps for the unlearned image and the retained image following the application of USGD present a significant shift in the model's attention. 
This result underscores that USGD erases a substantial amount of information linked to the unlearned image, including information that is also associated with other images in the same class. Consequently, this makes it challenging for an attack model to discern accurately which images were utilized during model training, and yet, there is a drop in model utility. This observation is consistent with the results presented in TABLE \ref{maintable}, where USGD outperforms our method in Unlearning Efficiency  while trailing behind in Model Utility.

As for Bad Teaching and Amnesiac unlearning methods, the results on the unlearned image differ significantly from the retrained model, indicating a substantial departure in the performance of the unlearned image from the desired gold model. Such discrepancies can potentially give rise to a ``Streisand effect'', leading to a more pronounced inadvertent disclosure of information about the unlearned image. This observation is also consistent with the high Unlearning Efficiency values associated with these two methods, as detailed in TABLE~\ref{maintable}.

Both Fisher and IAU yield results that resemble a subset of the retrained model results, suggesting their ability to approximate the true change in model parameters. The result from IAU follows a similar trend to the retrained model, displaying a slight downward shift.
Notably, it is intriguing to observe that the activation map on the retained image, as produced by the IAU method, closely mirrors the shape of the ones by the retrained model, in contrast to other methods where the activation maps remain unchanged or undergo shrinkage. This observation implies that \textit{the IAU method possesses the capability not only to eliminate the influence of unlearned images but also to effectively adjust model parameters, enabling it to align with the gold model characteristics on the retained data samples}.

\section{Complementary Experiments}
\label{comp}
In this section, we show more experiments to show the effectiveness of our IAU algorithm in various scenarios. Since the time overhead of the Fisher method even exceeds retraining, we no longer report its results.
\subsection{Tabular Data Experiments}
\label{app:ta}
Machine unlearning is particularly beneficial in situations pertaining to tabular data, such as medical or purchase records. In this regard, we present the findings of our experiments on the Purchase100 tabular dataset on the three-layer MLP model. As illustrated in TABLE \ref{tab:mlp}, our proposed approach IAU demonstrates noteworthy advantages in unlearning when compared to all baselines, thereby highlighting the practicality of our method.

\begin{table}[t]
\centering
\renewcommand{\arraystretch}{1.2}
\caption{Unlearning performance comparison with baselines for 5\% points unlearned randomly from the original training points on tabular dataset Purchase100 and model MLP. The optimal outcome is represented in bold typeface.}
\resizebox{\linewidth}{!}{
\begin{tabular}{c|c|cccc}
\hline
\multirow{2}{*}{Backbone} & \multirow{2}{*}{Strategy} & \multicolumn{4}{c}{Purchase100}   \\ \cline{3-6} 
                          &                           & MU$\downarrow$  & time$\downarrow$ & UE$\downarrow$    & Avg rank$\downarrow$ \\ \hline
\multirow{5}{*}{MLP}    & Retrain                     & 0    & 138  & 0     & -        \\
                          & USGD              & 0.50 & 3    & 7.94 & 0.7        \\
                          & Bad Teaching              & 7.02 & 8    & 24.30 & 2.7      \\
                          & Amnesiac Unlearning       & 2.49 & 44   & 13.19 & 2.3     \\
                          & Ours(IAU)                 & 0.21 & 2    & 8.19 & \textbf{0.3}        \\ \hline
\end{tabular}}
\label{tab:mlp}
\end{table}
\subsection{More Difficult Task}
\label{app:100}

To better demonstrate the effectiveness of the proposed method IAU, we extend the comparison of the model utility, required time, and unlearning efficacy achieved by the proposed method and the several baselines on a more difficult computer vision task.
we supplement the performance evaluation on the CIFAR-100 dataset and model VGG19, and the results are shown in TABLE \ref{tab:vgg} below. 
% Since Fisher requires a lot of time overhead, we omit its results. 
It is evident that our method still maintains comparable competitiveness, which again emphasizes the effectiveness of IAU.

\begin{table}[t]
\centering
\renewcommand{\arraystretch}{1.2}
\caption{Unlearning performance comparison with baselines for 5\% points unlearned randomly from the original training points on the CIFAR-100 dataset and model VGG19. The optimal outcome is represented in bold typeface.}
\resizebox{\linewidth}{!}{
\begin{tabular}{c|c|cccc}
\hline
\multirow{2}{*}{Backbone} & \multirow{2}{*}{Strategy} & \multicolumn{4}{c}{CIFAR100}   \\ \cline{3-6} 
                          &                           & MU$\downarrow$  & time$\downarrow$ & UE$\downarrow$    & Avg rank$\downarrow$ \\ \hline
\multirow{5}{*}{VGG19}    & Retrain                   & 0    & 747  & 0     & -        \\
                          & USGD              & 0.18 & 45   & 68.48 & 2        \\
                          & Bad Teaching              & 1.29 & 25   & 31.98 & 1.7      \\
                          & Amnesiac Unlearning       & 0.03 & 106  & 26.54 & 1.3      \\
                          & Ours(IAU)                 & 2.88 & 16   & 23.62 & \textbf{1}        \\ \hline
\end{tabular}}
\label{tab:vgg}
\end{table}
\subsection{Outlier Removal}
\label{app:out}

Machine unlearning becomes particularly challenging when faced with outliers. To better assess the proposed method IAU, we conducted an experiment on removing outliers.
We use the isolation forest \cite{liu2008isolation} to find outliers in the SVHN dataset. It reports 587 outliers out of 58606 total training data. We have supplemented the performance evaluation of removing those outliers of LeNet5, and the results are shown in TABLE \ref{tab:out} below. 
% Since Fisher requires a lot of time overhead, we omit its results. 
Compared with random forgetting (upper right part of TABLE \ref{maintable}), the performance improvement in deleting outliers is more significant, which confirms that the proposed method is particularly effective in dealing with outliers by the gradient-restriction method. 

\begin{table}[t]
\centering
\renewcommand{\arraystretch}{1.2}
\caption{Unlearning performance comparison with baselines for removing outliers task on SVHN dataset and model LeNet5. The optimal outcome is represented in bold typeface.}
\resizebox{\linewidth}{!}{
\begin{tabular}{c|c|cccc}
\hline
\multirow{2}{*}{Backbone} & \multirow{2}{*}{Strategy} & \multicolumn{4}{c}{SVHN}   \\ \cline{3-6} 
                          &                           & MU$\downarrow$  & time$\downarrow$ & UE$\downarrow$    & Avg rank$\downarrow$ \\ \hline
\multirow{5}{*}{LeNet5}    & Retrain                     & 0    & 538  & 0     & -        \\
                          & USGD              & 1.20 & 31    & 7.33 & 1.7        \\
                          & Bad Teaching              & 1.65 & 17    & 3.71 & 1.7    \\
                          & Amnesiac Unlearning       & 1.75 & 31   & 0.24 & 2     \\
                          & Ours(IAU)                 & 1.41 & 16    & 0.58 & \textbf{0.7}        \\ \hline
\end{tabular}}
\label{tab:out}
\end{table}

\subsection{Ablation Study}
\label{abl}
\begin{figure*}[htbp]
    \centering
    \subfloat[Model Utility]{\includegraphics[width= 0.5 \linewidth]{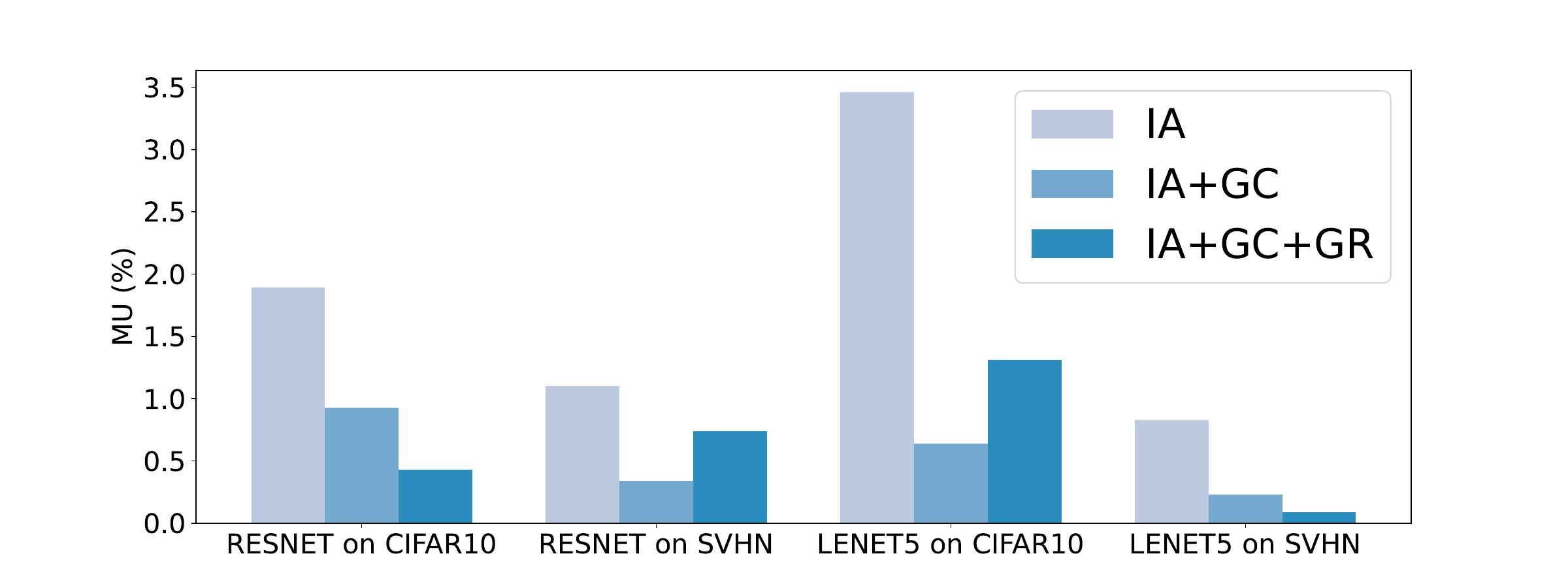}}
    \subfloat[Unlearning Efficacy]{\includegraphics[width= 0.5 \linewidth]{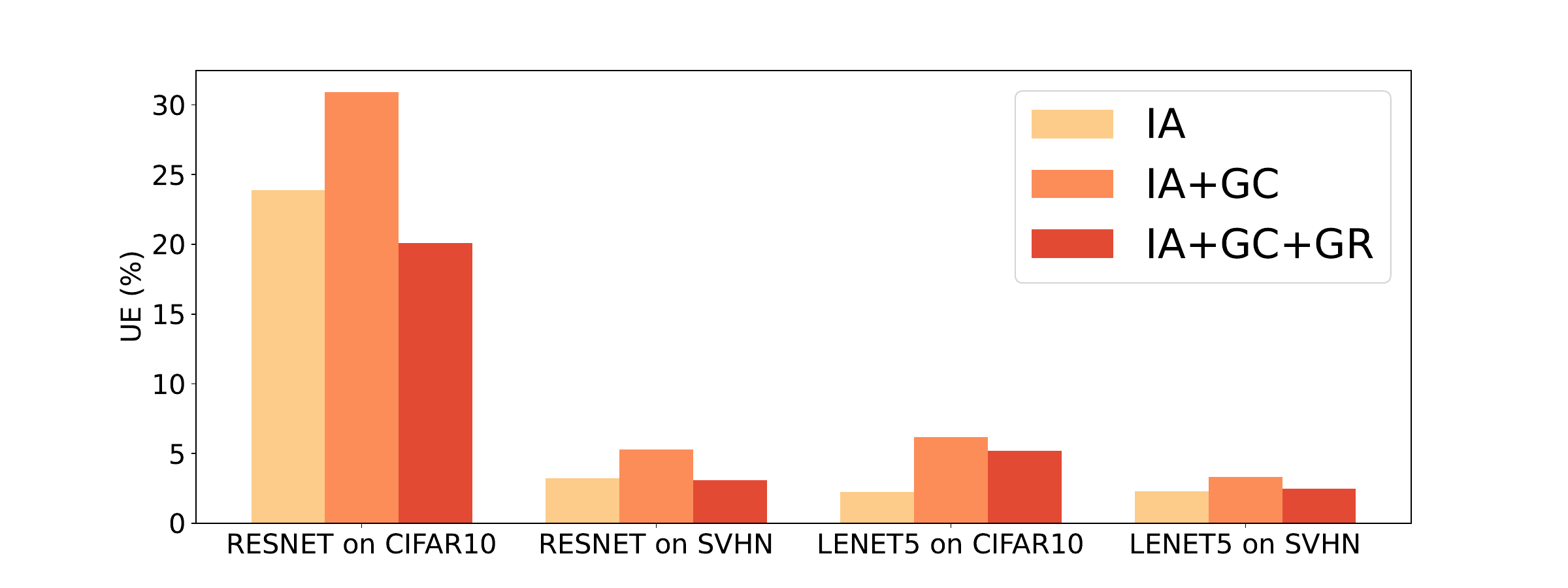}}
    \caption{\textbf{Ablation study} of Incremental Approximation (IA),  Gradient Correction (GC), and Gradient Restriction (GR) modules.}
    \label{fig:ablation}
\end{figure*}
In this section, we conduct an ablation study to investigate the role and interplay of the Incremental Approximation (IA),  Gradient Correction (GC) and Gradient Restriction (GR) modules in our method. The study has been structured to elucidate the distinct and collective influences of these components on the method's performance. The outcomes are presented in Fig~\ref{fig:ablation}. Model Utility (MU) and Unlearning Efficacy (UE) exhibit opposing trends, as discussed in Section \ref{subsec:hyper}. As a result, we can observe that the IA method leads to a bad in Model Utility with excellent Unlearning Efficacy, and IA+GC yields a significant enhancement in Model Utility but a decline in Unlearning Efficacy. However, when the IA+GC components are combined with GR (IA+GC+GR), there is an improvement in Unlearning Efficacy, while the Model Utility remains comparable to that of IA+GC. This implies that the IA+GC+GR combination achieves a more favorable trade-off between Model Utility and Unlearning Efficacy than other configurations.

\subsection{Effective of GR Loss}\label{subsec:gr}
In this section, we present the empirical evidence that supports the effectiveness of the proposed Gradient Restriction (GR). Our findings are based on a series of experiments designed to assess the impact of the proposed loss function (cf., Eq. \ref{eq: restriction}) on decreasing the required training epochs and minimizing the absolute value of the model gradient.
\begin{table}[t]
\centering
\renewcommand{\arraystretch}{1.2}
\caption{Number of training epochs required under early stopping for both the original loss and the Gradient Restriction (GR) loss. }
\begin{tabular}{ccccc}
\hline
\multicolumn{1}{c}{}     & \multicolumn{2}{c}{CIFAR10} & \multicolumn{2}{c}{SVHN} \\ \hline
\multirow{2}{*}{LENET5}   & Original Loss       & 13      & Original Loss     & 20     \\
                          & GR Loss           & 12      & GR Loss         & 18     \\ \hline
\multirow{2}{*}{ResNet18} & Original Loss       & 28      & Original Loss     & 12     \\
                          & GR Loss           & 25      & GR Loss         & 9      \\ \hline
\end{tabular}
\label{tab:epoch}
\end{table}

TABLE \ref{tab:epoch} shows the number of epochs required for model convergence before and after using GR loss. It can be seen that in all experimental settings, using GR loss requires fewer epochs. This verifies our discussion in Section \ref{sec: restriction} that penalizing high-valued gradients helps the model converge faster.

\begin{figure}[t]
    \centering

    \subfloat[CIFAR10]{\includegraphics[width= 0.5 \linewidth]{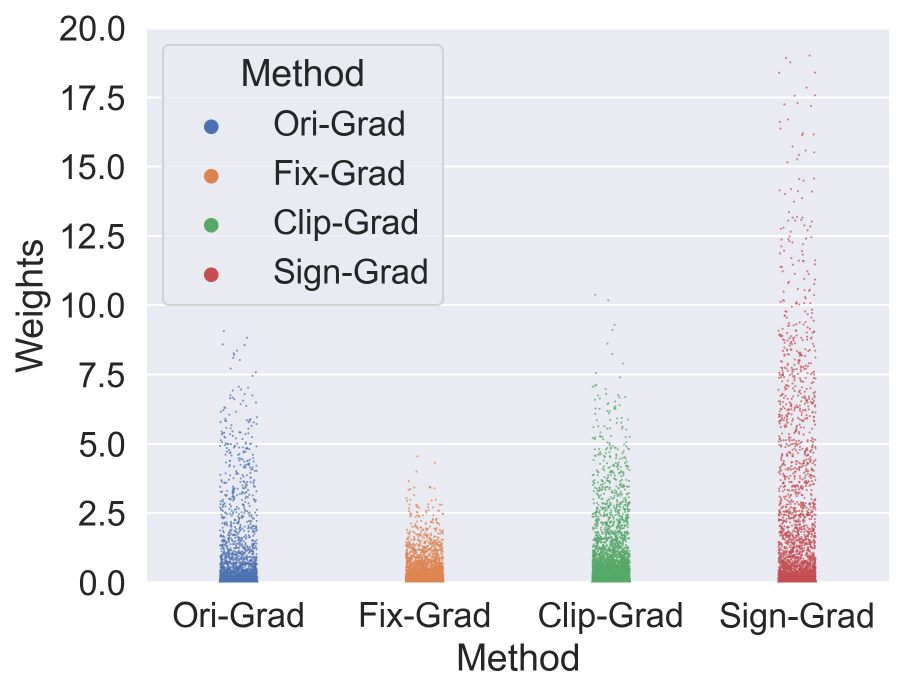}}
    \subfloat[SVHN]{\includegraphics[width= 0.5 \linewidth]{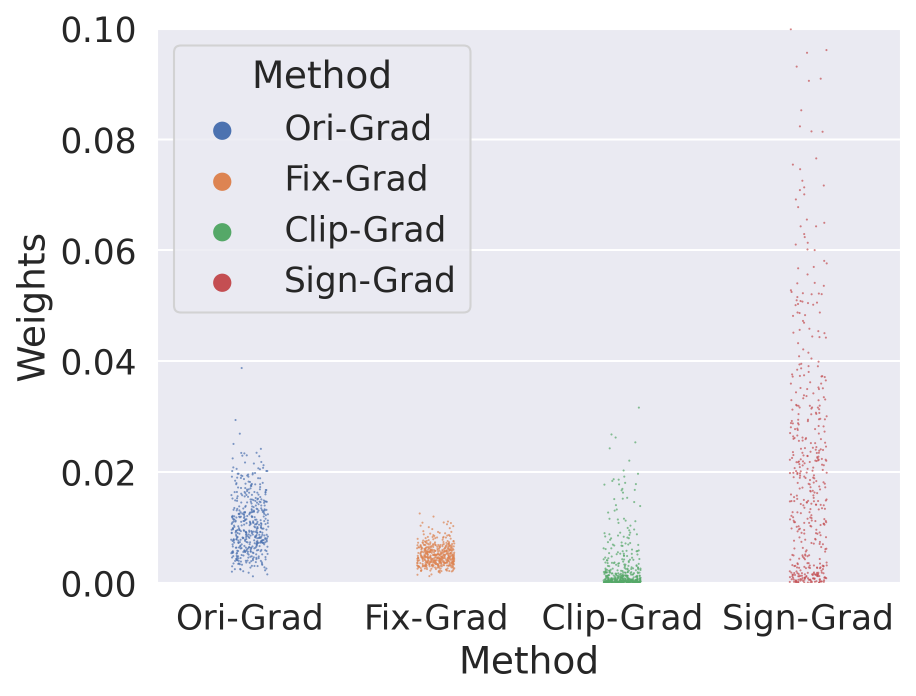}}
        \caption{The L2-norm of the gradients for ResNet18 is shown in four different scenarios after the model ceases training. The blue dots depict the original loss, the orange dots represent our Gradient Restriction loss, the green dots show gradient clipping, and the red dots are for SignSGD loss. These experiments were conducted on both CIFAR10 and SVHN datasets.}
    \label{fig:grad}
\end{figure}
\begin{figure}[t]

    \centering
    \subfloat[MU over $\alpha$]{\includegraphics[width= 0.5 \linewidth]{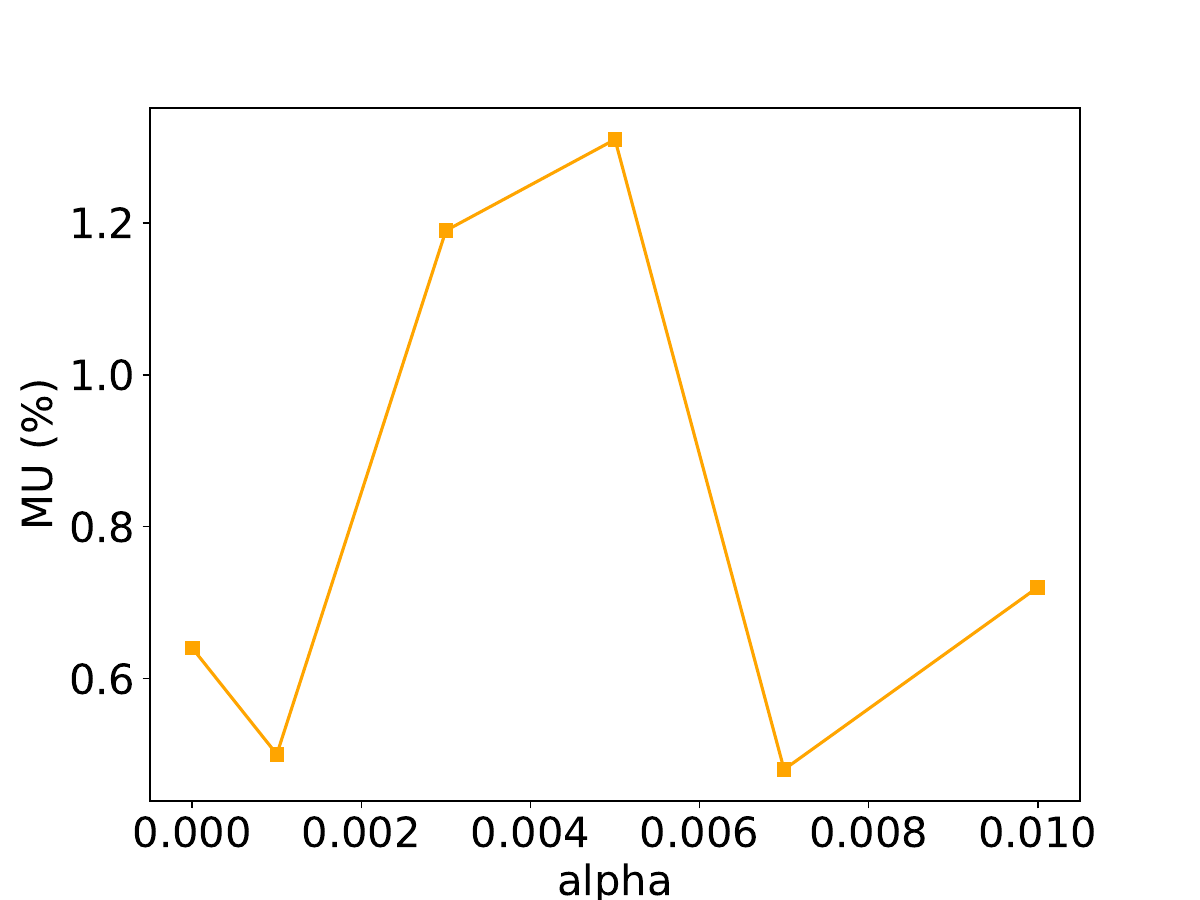}}
    \subfloat[UE over $\alpha$]{\includegraphics[width= 0.5 \linewidth]{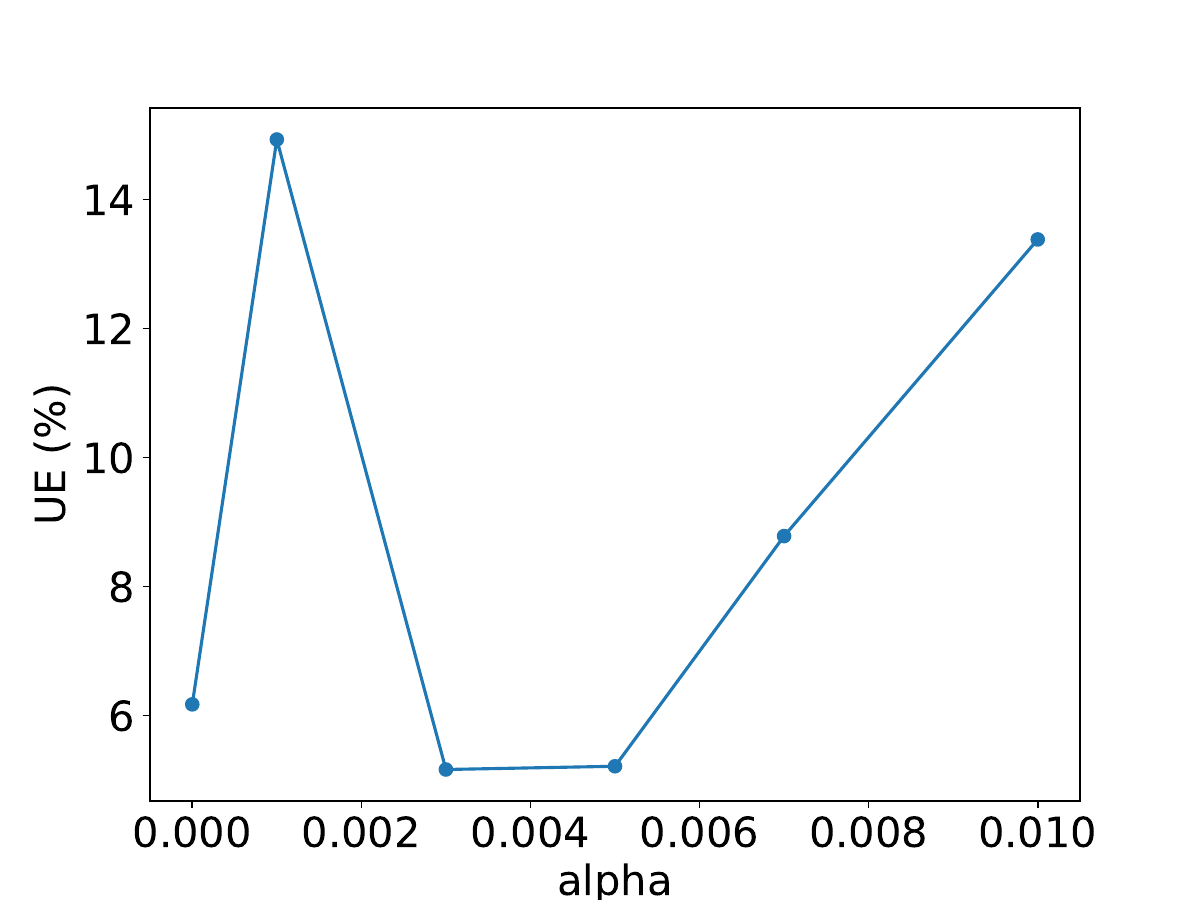}}

    \centering
    \subfloat[MU over $\rho$]{\includegraphics[width= 0.5 \linewidth]{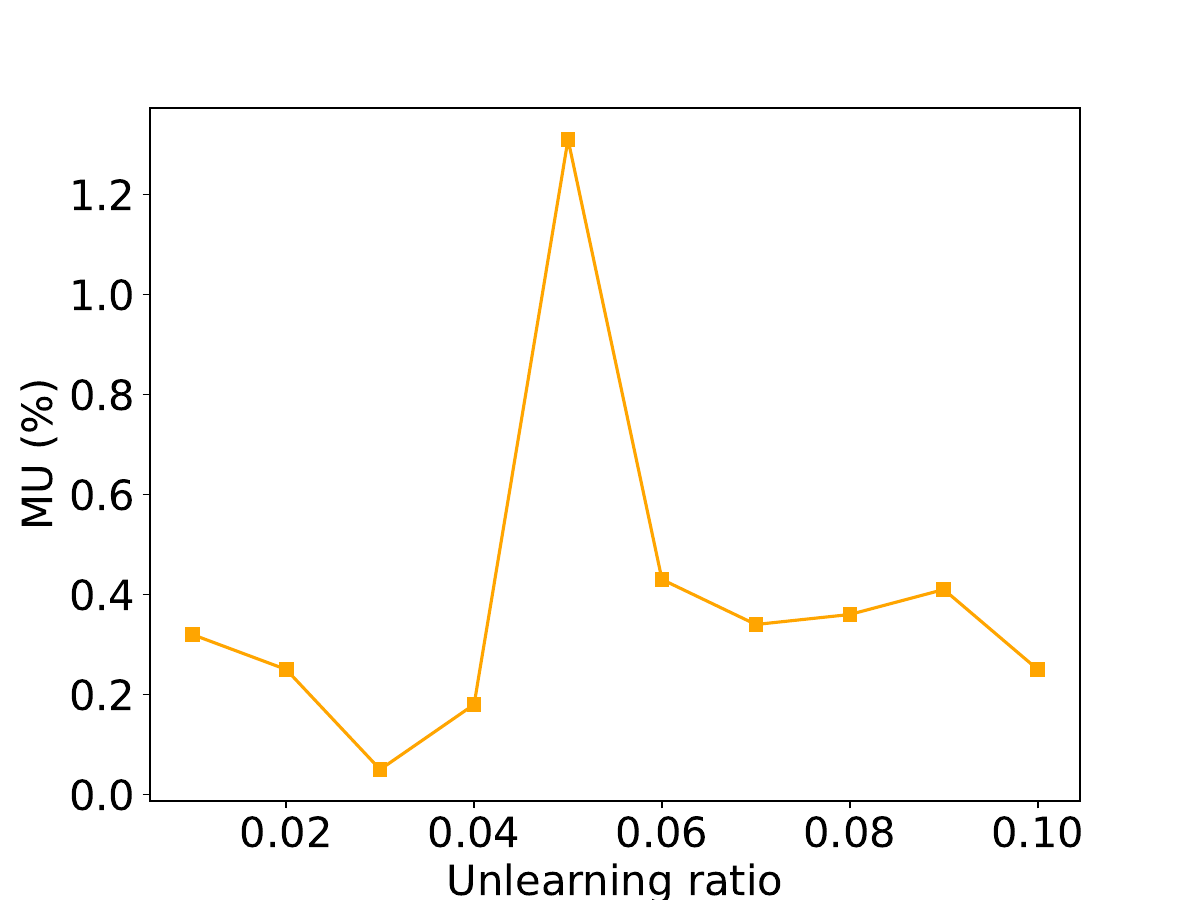}}
    \subfloat[UE over $\rho$]{\includegraphics[width= 0.5 \linewidth]{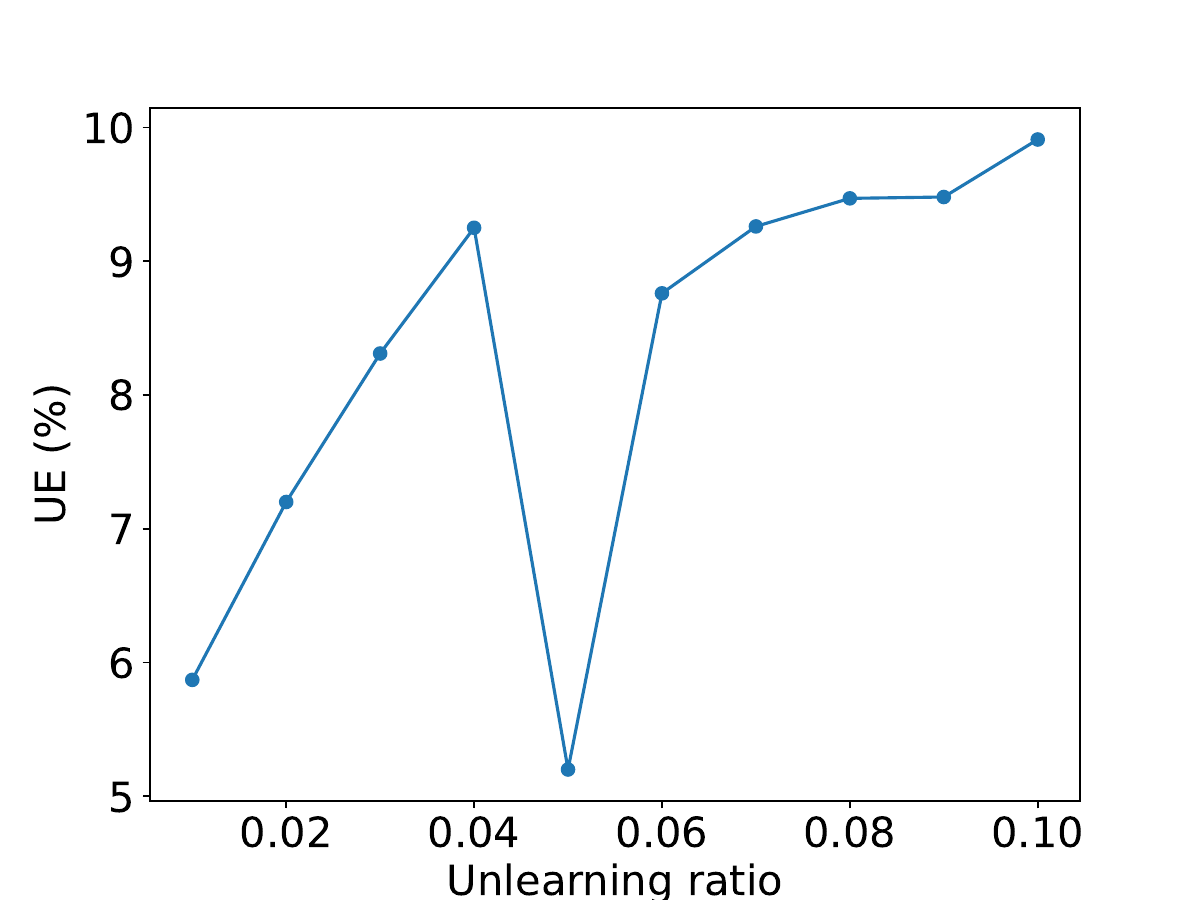}}
    \caption{Impact of parameter $\alpha$ in the GR loss function and  unlearning ratio $\rho$ on LENET5 model with CIFAR10 dataset. Both Model Utility (MU) and Unlearning Efficacy (UE) are ultra-small indicators.}
    \label{fig:hyper}
\end{figure}

% In order to showcase the effectiveness of our Gradient Restriction (GR) loss, we have conducted a comparison with gradient clipping \cite{clip} and SignSGD \cite{signsgd} methods. Fig. \ref{fig:grad} visually depicts the L2-norm distribution of model gradients for ResNet18 after convergence, clearly demonstrating the advantages of utilizing the GR loss. The results show that the GR loss results in a reduced distribution when compared to other methods, indicating that it effectively minimizes the L2-norm value of model gradients, enabling smoother and more controlled convergence during training.

In addition, to verify the reasonableness of the proposed GR loss, we compare it with existing gradient restriction methods, including Gradient Clip \cite{clip} and SignSGD \cite{signsgd}. Fig. \ref{fig:grad} visually depicts the L2-norm distribution of model gradients for ResNet18 after convergence. We can clearly find that Gradient Clip and SignSGD cannot limit the real gradient size, and even the gradient is larger than the origin (cf., Ori-Grad). Instead, Our method can limit the amplitude of the gradient. This underlines the rationale for the proposed GR loss.

Those experimental results provide robust evidence of the effectiveness of GR loss. This loss function not only enhances directional accuracy during training but also promotes smoother convergence by moderating gradients. These findings underscore the valuable impact of GR loss function in the context of efficient machine unlearning.
\subsection{Hyperparameter Study}\label{subsec:hyper}
Subsequently, we investigate hyperparameters, specifically examining unlearning tasks with parameter $\alpha$ in the GR loss function and different unlearning ratios denoted as $\rho$. These experiments are conducted on the CIFAR10 dataset using the LeNet5 model as our foundation.
Fig.\ref{fig:hyper} shows the experimental result with parameter $\alpha$ ranging from 0 to 0.10 and ratio $\rho$ range from 0.01 to 0.10.
We can observe that there exists an inverse relationship between Model Utility (MU) and Unlearning Efficacy (UE); high MU is associated with low UE, and vice versa. This observation aligns with our expectations since the unlearning process, based on the influence function, aims to approximate the change direction in model parameters rather than achieving the exact direction. Therefore, in order to enhance UE, the unlearning algorithm must induce more substantial changes in model parameters to effectively eradicate the influence of unlearned points, which, in turn, may adversely affect MU. Various values of $\alpha$ manifest different trade-offs between Model Utility (MU) and Unlearning Efficacy (UE). On average, the IAU method exhibits a limited sensitivity to changes in $\alpha$, as indicated by the relatively stable behavior of the y-axis. As $\rho$ increases, both Model Utility (MU) and Unlearning Efficacy (UE) deteriorate. This trend can be attributed to the inherently approximate nature of the unlearning process. With a higher ratio of unlearned data points, the results tend to become less precise, leading to increased vagueness in the outcomes.

\section{Conclusion}
This paper addressed the challenge of machine unlearning with minimal time overhead. We identified limitations in existing methods, particularly influence-based methods when dealing with large datasets and frequent unlearning demands. Drawing insights from cognitive science, we proposed an efficient unlearning method that approximates influence functions with high efficiency while preserving model utility. Our use of incremental learning in machine unlearning offers a novel perspective and has the potential to inspire future research. The empirical analysis demonstrated our method's efficiency in erasing learned information while maintaining model efficacy, surpassing current state-of-the-art methods in removal guarantee, unlearning efficiency, and comparable model utility.

% \section*{Acknowledgments}
% This should be a simple paragraph before the References to thank those individuals and institutions who have supported your work on this article.

% {\appendix[Proof of the Zonklar Equations]
% Use $\backslash${\tt{appendix}} if you have a single appendix:
% Do not use $\backslash${\tt{section}} anymore after $\backslash${\tt{appendix}}, only $\backslash${\tt{section*}}.
% If you have multiple appendixes use $\backslash${\tt{appendices}} then use $\backslash${\tt{section}} to start each appendix.
% You must declare a $\backslash${\tt{section}} before using any $\backslash${\tt{subsection}} or using $\backslash${\tt{label}} ($\backslash${\tt{appendices}} by itself
%  starts a section numbered zero.)}

\appendix
\section*{Proof of Theorem \ref{{theorem}[influence of adding a point]}}\label{proof}
% \addcontentsline{toc}{section}
We define $R(\theta)$ as the empirical risk of a model $h$:
$$
R(\theta) \stackrel{\text { def }}{=} \frac{1}{n} \sum_{i=1}^{n} \ell(z_i, \theta). $$
Empirical risk minimization(ERM) is the method of finding the minimizer of $R(\theta)$, which we call 
$$
\theta^* \stackrel{\text { def }}{=} \arg \min _\theta \frac{1}{{n}} \sum_{i=1}^{n} \ell(z_{i}, \theta).$$
We assume that $R$ is strictly twice-differentiable and convex; thus we know that
$$
H_{{\theta}^*} \stackrel{\text { def }}{=} \nabla_\theta^2 R({\theta}^*)=\frac{1}{n} \sum_{i=1}^{n} \nabla_\theta^2 \ell(z_{i}, \theta^*)$$
exists and is positive definite.
After adding a point $z_+$ and up-weighting it by an infinitesimal amount $\mu$ on original model $\theta^*$, the new model $\theta_{\{z_+\}}^\mu$ is defined as
% \begin{equation*}
\begin{align*}
\theta_{\{z_+\}}^\mu &\stackrel{\text { def }}{=}
% \arg \min _\theta \frac{1}{n+1}(\sum_{i=1}^{n} \ell(h_\theta(z_i))+ \ell(h_\theta(z_+)))  \\
% & =\arg \min _\theta \frac{1}{n+1} \sum_{i=1}^{n} \ell(h_\theta(z_i))+\frac{1}{n+1} \ell(h_\theta(z_+)) \\
\arg \min _\theta \frac{n}{n+1}\left(\frac{1}{n} \sum_{i=1}^{n} \ell(z_i,\theta)\right)+\mu \ell(z_+,\theta) \\
& =\arg \min _\theta \frac{n}{n+1}R(\theta)+\mu\ell(z_+,\theta). \\
\end{align*}
We define the new empirical risk minimizer as
$$
 {\theta}^*_{\{z_+\}} \stackrel{\text { def }}{=} \operatorname{argmin}_{\theta} \frac{1}{{n+1}} \left(\sum_{{i}=1}^{{n}} {\ell}({z}_{{i}}, \theta)+{\ell}(z_+,\theta)\right). $$
% \end{equation*}
$\theta_{\{z_+\}}^\mu$ is the minimizer of $\frac{n}{n+1}R(\theta)+\mu\ell(z_+, \theta)$, then we have
$$
\frac{n}{n+1}\nabla_\theta R(\theta_{\{z_+\}}^\mu)+\mu\nabla_\theta \ell(z_+,\theta_{\{z_+\}}^\mu)=0.$$
Next, since $\theta_{\{z_+\}}^* \rightarrow \theta^*$ as $\mu \rightarrow 0$, we can perform a Taylor expansion:
\begin{align*}
0 \approx &\frac{ {n}}{ {n}+1}\left[\nabla_\theta  {R}(\theta^*)+\nabla_\theta^2  {R}(\theta^*)(\theta_{\{z_+\}}^\mu-\theta^*)\right]+\\
&\mu \nabla_\theta \ell(h_{\theta^*}(z_+))+\mu \nabla_\theta^2  \ell(z_+,\theta^*)(\theta_{\{z_+\}}^\mu-\theta^*)
\end{align*}
where we have dropped $o(||\Delta_\mu||)$ terms.
Defining the parameter change $\Delta_{\mu}\stackrel{}{=}\theta_{\{z_+\}}^\mu-{\theta}^*$, we have:
\begin{align*}
0 \approx &\frac{ {n}}{ {n}+1}\left[\nabla_\theta  {R}(\theta^*)+\nabla_\theta^2  {R}(\theta^*)\Delta_{\mu}\right]+\\
&\mu \nabla_\theta \ell(z_+, \theta^*)+\mu \nabla_\theta^2  \ell(z_+,\theta^*)\Delta_{\mu}.
\end{align*}
Arrange the above formula to get
\begin{align*}
    0 \approx &\frac{ {n}}{ {n}+1} \nabla_\theta  {R}( \theta^*)+\mu \nabla_\theta \ell(z_+, \theta^*)+\\
    &\left[\frac{ {n}}{ {n}+1} \nabla_\theta^2  {R}( \theta^*)+\mu \nabla_\theta^2  \ell(z_+, \theta^*)\right] \Delta_\mu 
\end{align*}
 Solving for $\Delta_\mu$, we get:
\begin{align*}
\Delta_\mu \approx&-\left[\frac{ {n}}{ {n}+1} \nabla_\theta^2  {R}( \theta^*)+\mu \nabla_\theta^2  \ell(z_+, \theta^*)\right]^{-1}\\
&\left[\frac{ {n}}{ {n}+1} \nabla_\theta  {R}( \theta^*)+\mu \nabla_\theta \ell(z_+, \theta^*)\right]
% \approx& -\nabla_\theta^2  {R}( \theta^*)^{-1} \left[\frac{ {n}}{ {n}+1} \nabla_\theta  {R}( \theta^*)+\mu \nabla_\theta \ell( {z}_{+},  \theta^*)\right]
\end{align*}
Since $\theta^*$ is the minimizer of $R$, we have $\nabla_\theta  {R}( {\theta}^*)=0$. 
Only keeping $O(\mu)$ terms, we get
\begin{align*}
\Delta_\mu \approx&-\left[\frac{ {n}}{ {n}+1} \nabla_\theta^2  {R}( \theta^*)\right]^{-1} \mu \nabla_\theta \ell(z_+,\theta^*) .
\end{align*}
Thus we can have
\begin{align*}
    \left.\frac{d\theta_{\{z_+\}}^\mu}{d \mu}\right|_{\mu=0}=\left.\frac{d\Delta_\mu}{d \mu}\right|_{\mu=0}=-\frac{ {n+1}}{ {n}}H_{\theta^*}^{-1} \nabla_\theta \ell(z_+, \theta^*)
\end{align*}
This yields
$$
\theta_{\{z_+\}}^\mu\approx \theta^*-\mu\frac{ {n+1}}{ {n}}H_{\theta^*}^{-1} \nabla_\theta \ell(z_+, \theta^*).
$$
As adding a point $z_+$ is equal to up-weighting it by $\mu=\frac{1}{n+1}$, we can get approximation of parameter change 
$$\theta_{\{z_+\}}^* \approx \theta^*-\frac{1}{n}H_{\theta^*}^{-1} \nabla_\theta \ell(z_+, \theta^*).
$$

\normalem
\bibliographystyle{IEEEtran}
\bibliography{1example_paper}

\newpage

\section{Biography Section}
 
\vspace{11pt}

\vfill

\end{document}